\PassOptionsToPackage{numbers,sort&compress}{natbib}
\documentclass{article}

\usepackage[preprint]{corl_2026} 

\usepackage{multicol}
\usepackage{amsmath}
\usepackage{amssymb}
\usepackage{graphicx}
\usepackage{subcaption}
\usepackage{booktabs}
\usepackage[font=small]{caption}
\usepackage{mathtools}
\usepackage{enumitem}
\usepackage{wrapfig}

\title{Where to Touch, How to Contact: A Hierarchical RL--MPC Framework for Geometry-Aware Sim-to-Real Manipulation}
\author{
  Zhixian Xie\\
  Arizona State University\\
  \And
  Yu Xiang\\
  University of Texas at Dallas\\ 
  \AND
  Michael Posa\\
  University of Pennsylvania\\ 
  \And
  Wanxin Jin\\
  Arizona State University\\
}

\begin{document}
\maketitle
\vspace{-30pt}

\begin{abstract}
A key challenge in contact-rich dexterous manipulation is the need to jointly reason over global geometry and  nonsmooth contact dynamics. End-to-end  policies bypass this complexity, but often require large amounts of data and transfer poorly from simulation to reality. We address the limitations with a simple insight: dexterous manipulation is inherently hierarchical—at a high level, a robot decides where to touch (geometry); at a low level it determines how to move the object through contact dynamics. Building on this insight, we propose a hierarchical RL--MPC framework in which a high-level reinforcement learning (RL) policy predicts a contact intention, a novel object-centric interface that specifies (i) an object-surface contact location and (ii) a post-contact object subgoal pose. Conditioned on the contact intention, a low-level contact-implicit model  predictive control (MPC) optimizes local contact modes and real-time (re)plans through contact dynamics to generate robot actions that robustly move the object toward each subgoal. We evaluate the framework on non-prehensile tasks, including geometry-generalized pushing across diverse object shapes, pivoting/flipping-based object reorientation, and environment-assisted object repositioning. It achieves high success rate with substantially reduced data (${\sim}40\times$ fewer RL decision steps and ${\sim}2\times$ fewer control steps in the T-pushing comparison), highly robust performance, and zero-shot sim-to-real transfer. \href{https://zhi-xian-xie.github.io/contact_intention_website/}{Project Link}.
\end{abstract}



\section{Introduction}
\vspace{-10pt}
Dexterous manipulation is a hallmark skill of intelligent robots. Recent research has made significant progress by training robot policies either from  demonstrations, e.g., generative policy imitation  \cite{zhao2023learning,chi2025diffusion} and VLA model \cite{black2410pi0, kim2024openvla, figure2024helix}, or from robot experience, e.g., reinforcement learning \cite{qi2023hand,yin2023rotating}. These approaches typically learn an end-to-end policy from robot observations to low-level motor commands. While end-to-end methods can achieve impressive performance, they often face practical challenges, such as data inefficiency and  sim-to-real gap.

Yet, human manipulation does not necessarily use a monolithic end-to-end motor policy \cite{merel2019hierarchical,grafton2010cognitive}. For instance, when picking up a hammer to strike a nail, we do not directly infer low-level finger–joint commands from raw perception. Instead, we reason hierarchically: at a high level, we decide \emph{where} to contact (e.g., grasping the handle) and \emph{what} object-level outcome to achieve  (e.g., aligning the head with the nail); at low level, we determine \emph{how} to realize this plan based on our intuitive  physics, i.e., the anticipated outcomes of contact strategies of rolling, sliding, sticking, and  disengagement. This hierarchical decomposition enables humans to combine  geometric understanding and physics intuition, yielding efficient and robust manipulation control.

Building on this intuition, we propose a hierarchical reasoning framework for efficient, geometry-aware dexterous manipulation. We target contact-rich tasks whose semantic goals may be simple, such as pushing or reorienting an object, but whose execution requires coupled reasoning over geometry and contact dynamics: where to make and switch contact, and how to exploit rolling, pivoting, sticking, sliding, and disengagement. The framework consists of:

\begin{itemize}[leftmargin=*, itemsep=0pt, topsep=2pt, parsep=0pt, partopsep=0pt]
\item \textbf{High level:  geometric reasoning.}
Given a scene observation, the high-level RL policy predicts a \emph{Contact Intention}: a high-level description of contact locations on the object surface and a post-contact object subgoal, specifying where to interact and what object-level motion to induce.


\item \textbf{Low level: contact dynamics reasoning.}
Conditioned on the contact intention, the low-level contact-implicit predictive control optimizes robot action commands through contact dynamics, synthesizing local contact modes such as rolling, sticking, sliding, and disengagement to robustly drive the object toward the subgoal.
\end{itemize}
\vspace{-5pt}
\noindent
The two levels forms a closed-loop hierarchical policy that offers the following benefits:
\emph{(I) Significant data efficiency compared to training end-to-end RL policies.}
By delegating non-smooth contact dynamics reasoning to MPC, the policy focuses on geometry and kinematics reasoning, enabling efficient learning of long-horizon contact plan.
\emph{(II) Better global contact planning than pure model-based optimization.}
MPC and trajectory optimization \cite{jin2024complementarity,posa2014direct} can reason over contact dynamics, but they are often short-horizon and prone to local optima. The high-level RL policy complements them by selecting geometry-aware contact locations and object subgoals, guiding the low-level optimizer toward more effective long-horizon contact plans.
\emph{(III) Enabling sim-to-real transfer.}
The contact-intention interface decouples geometry reasoning from low-level contact dynamics, allowing RL policy to be trained in simulation where geometry/kinematics transfer more reliably than contact dynamics. It makes the learned higher-level RL policy  more robust for sim-to-real deployment than an end-to-end policies.
We highlight our following contributions:

(1) We propose a hierarchical geometry--dynamics reasoning framework for efficient, geometry-aware dexterous manipulation. A higher-level RL policy predicts \emph{contact intentions}---object-surface contact locations and post-contact object subgoals---while a low-level contact-implicit MPC realizes them through physically grounded contact (re)planning and control.

(2) We design a generalizable high-level RL policy for contact-intention prediction. The policy uses a tri-component object-centric observation that encodes object surface geometry, task-progress flow, and environment-contact distances, enabling geometry-aware exploration across diverse object shapes. Because the policy predicts object-level contact intentions rather than robot actions, it can be trained with an object-level model and deployed with robot-specific MPC.


(3) We validate the framework in simulation and real-world experiments on diverse contact-rich tasks, including geometry-generalized pushing, pivoting/flipping-based object reorientation, and environment-assisted object repositioning. It achieves high data efficiency, robust success rates, and zero-shot sim-to-real transfer compared with end-to-end RL and model-based baselines.

\vspace{-8pt}
\section{Related Works}
\vspace{-8pt}
\paragraph{RL-MPC Hierarchy in Robot Manipulation}
The use of RL–MPC architectures is extensive in robotic locomotion \cite{yang2022fast,zhang2022model,chen2024learning,jia2024rl,hu2025lno}, where the primary objective is to achieve agile, adaptive, and diverse motor behaviors by combining learned high-level decision making with model-based control. This paradigm has more recently been used in robotic manipulation. In \cite{shin2019autonomous, liu2025autonomous}, RL is employed to infer the latent dynamics of deformable objects, which are subsequently incorporated into MPC or adaptive control for accurate manipulation.  \cite{omer2021model, lopes2025model} leverage RL to assist with learning contact-rich dynamics models and to provide a policy warm-start for sampling-based MPC solvers. In another line of research \cite{bing2023safety, zhuang2025rm}, RL predicts intermediate end-effector goals for a low-level MPC tracking controller,  ensuring feasibility and safety during contact-rich tasks. Unlike prior methods that use RL mainly for auxiliary guidance or trajectory prediction, in our proposed hierarchy, RL is used for geometric reasoning, while MPC handles contact-dynamics reasoning. To synergizes both levels, we propose a novel \emph{contact intention} interface, which is an abstract action space of RL policy.

\vspace{-8pt}
\paragraph{Non-Prehensile Manipulation} Since the proposed method is primarily applied to non-prehensile manipulation, we review the relevant prior work below.
Non-prehensile manipulation \cite{mason1986mechanics} is defined as manipulating objects without explicitly grasping them. Classic work in non-prehensile manipulation studied the mechanics and control of the manipulator-object motion primitives \cite{lynch1996stable, akella1998posing}. However, motion primitives
face scalability issue in the open-ended world. Recent methods solve the problem through contact-implicit planning \cite{mordatch2012contact, moura2022non, yi2023precise, wang2022contact, posa2014direct, cheng2023enhancing,kurtz2023inverse,aydinoglu2024consensus, yang2024dynamic, bui2025push}, which allows a robot to automatically discover contact modes through optimization through contact dynamics. Learning methods have also been used in non-prehensile manipulation, such as the work based on reinforcement learning  \cite{zhou2023hacman, cho2024corn,cho2025hierarchical,li2025pin,lin2025sim}, imitation learning \cite{chi2025diffusion, saigusa2022imitation, zhu2023viola, rouxel2024flow, tosun2019pixels, wang2025hierarchical} or both \cite{sun2022integrating}. 
Recent work of PushAnything~\cite{bui2025push}, HiDex~\cite{cheng2023enhancing}, and HDP~\cite{wang2025hierarchical} is also related to our work, however, PushAnything samples contact locations and evaluates them via MPC; HiDex uses MCTS for contact selection followed by path planning; and HDP employs a diffusion-based hierarchy for contact prediction and trajectory generation. Those methods differ from ours in the design of control hieriarchy.
Specifically, we emphasize the distinction from HACMan~\cite{zhou2023hacman}, which also decomposes manipulation into \emph{where} to touch and \emph{how} to contact. However, HACMan uses a flat RL policy to jointly learn both components, requiring the policy to implicitly handle contact dynamics. In contrast, our method offloads contact-dynamics reasoning to contact-implicit MPC, allowing the RL policy to focus on geometric reasoning. This decomposition yields significantly better data efficiency and higher manipulation performance.

\vspace{-8pt}
\paragraph{Robot Manipulation Based on Affordance} Our work is also related to affordance-based manipulation, where affordance describes the contact location. However, most of the research has focused grasping tasks \cite{hermans2011affordance, geng2022end, xu2021affordance, chu2019toward, chu2019learning, jiang2021synergies, li2025learning}. Recently, foundation models have further been used to infer affordance for robotic tasks \cite{huang2023grounded, ji2025robobrain, tang2025roboafford, li2024manipllm, song2025maniplvm, tong2024oval}.
Affordance prediction only indicates \emph{where} to make contact and lacks guidance on \emph{how} to act after contact. Our method addresses  dynamically predicting contact intentions online, allowing the policy to switch contact locations as the object state evolves, and coupling them with a physics-based planner to generate adaptive, physically grounded non-prehensile motions.
\begin{figure}[t]
    \centering
    \includegraphics[width=0.8\linewidth]{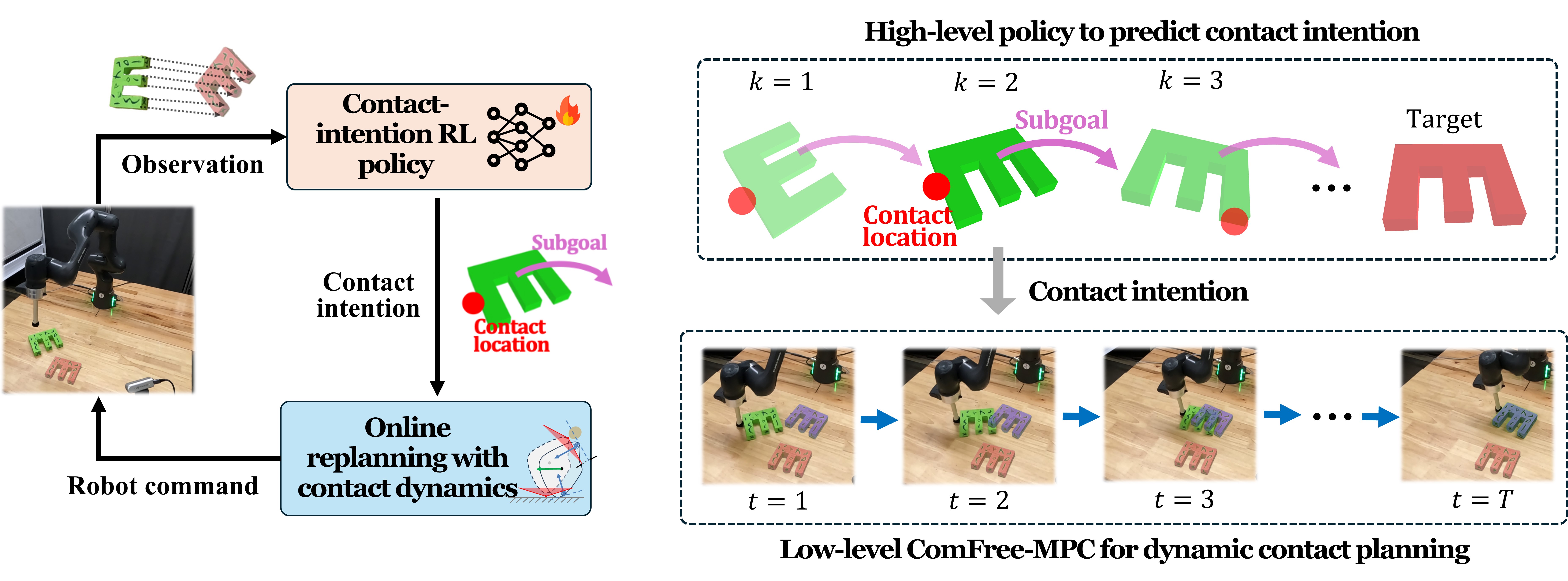}
    \caption{Overview of the propoed RL-MPC hierarchical framework}
    \vspace{-15pt}
    \label{fig.holistic}
\end{figure}
\vspace{-8pt}
\section{Contact Intention and Hierarchical RL-MPC Framework}
\vspace{-8pt}
\subsection{Contact Intention for Contact-Rich Manipulation}
\vspace{-8pt}
\label{sec.intention}
Consider $n$ \emph{active} contact points on a robot that can be  actuated to interact with an object. For example, $n=1$ for a robot arm  with a stick-like end-effector, and $n=4$ for the fingertips of a four-fingered robotic hand. We distinguish these active contact points from passive ones that arise without deliberate  interaction, e.g.,  object–environment (or other robot links) contacts.

The goal of  manipulation  is for the robot to purposefully engage its $n$ active contact points to interact with an object through contact and drive the object to a target pose $\boldsymbol{\bar{q}}^{\text{target}}$. For this, 
we define \textbf{Contact Intention} as a high-level action:
\vspace{-1pt}
\begin{equation}\label{equ.ci_space}
    \mathbf{I} = (\boldsymbol{\bar{c}}^1, \boldsymbol{\bar{c}}^2, \ldots, \boldsymbol{\bar{c}}^n, \, \boldsymbol{\bar{q}}^{\text{obj}}) \in \mathcal{G}^n \times \mathrm{SE}(3),
    \vspace{-0pt}
\end{equation}
Here, $(\boldsymbol{\bar{c}}^1, \boldsymbol{\bar{c}}^2, \ldots, \boldsymbol{\bar{c}}^n)$ are the corresponding contact locations on the object surface $\mathcal{G}$.  Each  location $\boldsymbol{\bar{c}}^i\in\mathcal{G}$, $i=1,..,n$, specifies an \emph{interaction anchor} on the object surface  for an active contact point $i$ to engage with. $\boldsymbol{\bar{q}}^{\text{obj}} \in \mathrm{SE}(3)$ is the  post-contact subgoal pose of the object. It specifies the desired object-level outcome for the robot-object interaction at the suggested contact locations.  The subgoal may not coincide with the final target  $\boldsymbol{\bar{q}}^{\text{target}}$; it may be an intermediate object pose that provides meaningful progress toward $\boldsymbol{\bar{q}}^{\text{target}}$. In this paper, we only consider non-prehensile manipulation using a stick-like end-effector attached to a robot arm. Thus,  $n=1$. 
\vspace{-8pt}
\subsection{Overview of RL-MPC Hierarchical Framework}
\vspace{-8pt}
\label{sec.overview}
Our proposed RL-MPC hierarchical framework is shown in  Fig. \ref{fig.holistic}. It consists of a high-level  policy  for object-centric  geometry reasoning  and a low-level policy for  contact dynamics reasoning. 
The high-level policy, trained using RL, predicts contact intention, which parameterizes the  low-level   policy, which online optimizes the robot action through local contact-implicit MPC.

The high-level geometry reasoning policy predicts  contact intention, we formulate it as a POMDP, where the action space $\mathcal{A}$ is the whole contact intention space $\mathcal{A}=\mathcal{G}^n\times \mathrm{SE}(3)$. A high-level policy  $\pi$ maps observations to contact intentions. At each decision step $k$, the policy gets an observation $\boldsymbol{o}_k$ and predicts contact intention $\mathbf{I}_k=\pi(\boldsymbol{o}_k)\in\mathcal{A}$. The predicted  $\mathbf{I}_k$  will parameterize  the low-level MPC for  contact dynamics reasoning. We train $\pi$  in simulation, which will be detailed in Section \ref{sec.experiment}.

Given a contact intention $\mathbf{I}_k=(\boldsymbol{\bar{c}}^1_k, \boldsymbol{\bar{c}}^2_k, ..., \boldsymbol{\bar{c}}^n_k, \boldsymbol{\bar{q}}^{\text{obj}}_k)$ from the high-level, we formulate the   low-level contact-implicit dynamics planning as a complementarity-free model predictive control (ComFree-MPC)   \cite{jin2024complementarity}. Specifically,  ComFree-MPC reasons about the rich local contact strategies at the suggested contact location $(\boldsymbol{\bar{c}}^1_k, \boldsymbol{\bar{c}}^2_k, ..., \boldsymbol{\bar{c}}^n_k)$ to move the object towards the subgoal $\boldsymbol{\bar{q}}^{\text{obj}}_k$. At each control  step $t$,  ComFree-MPC  plans  $H$ steps ahead through complementarity-free contact dynamics and finds the best robot action sequence for moving the object to $\boldsymbol{\bar{q}}^{\text{obj}}_k$.
The optimization is written as
{\small
\begin{equation}
\begin{aligned}
    \min_{\boldsymbol{u}_{0:H{-}1}}  \quad & \sum\nolimits_{t=0}^{H-1} \sum\nolimits_i L(\mathbf{p}_t^{{\mathrm{ee}},i}, \, \boldsymbol{\bar{c}}^{\,i}_k)+V(\boldsymbol{q}^{\text{obj}}_H,\boldsymbol{\bar{q}}^{\text{obj}}_k) \\
    \text{s.t.} \quad & \boldsymbol{x}_{t+1} = \texttt{ComFree}(\boldsymbol{x}_{t}, \boldsymbol{u}_{t}), t=0,1,...,H{-}1 \quad \text{Given} \quad \boldsymbol{x}_0{=}\boldsymbol{x}^{\text{env}}
    \label{equ.mpc}
\end{aligned}
\end{equation}}

Here, $\boldsymbol{x}_{0:H}$ is the predicted state of the manipulation system (robot and object). $\boldsymbol{u}_{0:H-1}$ is the robot action input sequence. The MPC objective  includes: (1) a running cost $L$ which encourages the robot's active contact points $(\mathbf{p}_t^{{\mathrm{ee}},i})_{i=1}^{n}$ to get close to the suggested locations $(\boldsymbol{\bar{c}}^i_k)_{i=1}^{n}$.  (2) The terminal cost $V$ encourage the robot  to move the object $\boldsymbol{q}^{\text{obj}}_H$ to the  subgoal $\boldsymbol{\bar{q}}^{\text{obj}}$.
\texttt{ComFree} is the complementarity-free contact physics model \cite{jin2024complementarity, borse2026comfree}, enabling fast receding horizon optimization.
$\boldsymbol{x}_0$ is set as the system’s current environment state at each MPC call. After solving the optimization, only the first control input  is sent  as the action command to the robot. See details will in Section \ref{sec.MPC}.

\emph{Multi-Rate  RL-MPC Coupling:} 
The ComFree-MPC  executes for $T$ environment steps while the contact intention from the high-level policy remains fixed. Below, we use $k$ to denote the high-level decision step for updating contact intentions and $t$ to denote the low-level MPC control step.
\vspace{-14pt}
\section{RL for  Geometry Reasoning}\label{sec:observation}
\vspace{-8pt}
\vspace{-0pt}
We use RL to train the  high-level policy to take in an observation  and output a contact intention. To enable effective training, we design the observation and action spaces in this section,  with detailed policy architecture shown in Appendix \ref{appendix:backbone}.
\begin{figure}[!htbp]
\vspace{-10pt}
    \centering
    \includegraphics[width=0.9\linewidth]{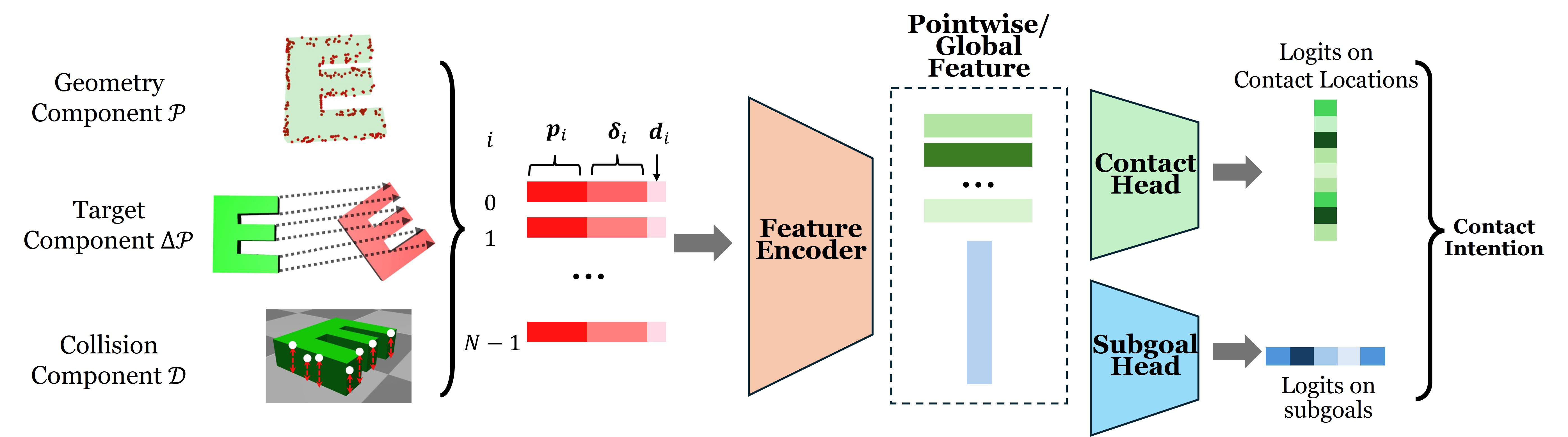}
    \caption{Observation and network design for the high-level policy. Left: geometric, target, and environment collision components of the RL observation. Right: policy network architecture, where per-point features are encoded into local or global representations, followed by prediction heads for action logits.}
    \vspace{-10pt}
    \label{fig:RL_obs_nets}
\end{figure}

\vspace{-5pt}
\paragraph{RL Observation Space Design}

As  in Fig.~\ref{fig:RL_obs_nets} (left), we design the high-level observation  as a \emph{tri-component, object-canonical-frame} representation  $\boldsymbol{o}_k = \big(\mathcal{P},\, \Delta \mathcal{P}_k,\, \mathcal{D}_k\big)$, where $\mathcal{P}=\{\boldsymbol{p}_i\}_{i=0}^{N-1}$ is the keypoint set on object surface, which approximates the object geometry and provides discrete candidates for contact-location selection. The target component is the object-frame keypoint goal flow 
$\Delta\mathcal{P}_k=\{\boldsymbol{\delta}_{i,k}\}_{i=0}^{N-1}$, which encodes the displacement of each keypoint from the current object pose toward the target pose. The collision component is the set of per-keypoint clearance distances 
$\mathcal{D}_k=\{d_{i,k}\}_{i=0}^{N-1}$, which indicates contact feasibility and helps the policy avoid environment-blocked contact locations. All those components are in the object frame, because it allows the RL policy to reason about object-centric geometry, task progress, and \emph{environment contact}  without ``rediscovering the object-pose-dependent transformation". We provide the detailed definition in Appendix \ref{appendix:obs}, and provide ablation study for our above design in Section \ref{sec.experiment}.

\vspace{-8pt}
\paragraph{RL Action  Design for Object Subgoal}
\label{sec.rl_action}
The RL policy (Fig.~\ref{fig:RL_obs_nets} (right)) receives the observation $\boldsymbol{o}$ and selects a contact intention $\mathbf{I}=(\boldsymbol{\bar{c}}^1, \boldsymbol{\bar{c}}^2, ..., \boldsymbol{\bar{c}}^n, \boldsymbol{\bar{q}}^{\text{obj}})$. 
Here, each contact location  is selected from the keypoint set $\mathcal{P}$, i.e., $\boldsymbol{\bar{c}}^i\in \mathcal{P}$, $i=1,2,..., n$. To reduce the complexity of RL policy searching over the entire  $\mathrm{SE}(3)$  for object pose $\boldsymbol{\bar{q}}^{\text{obj}}$, in this work we propose two surrogate representations.

\emph{Option 1: Subgoal re-parametrization using MPC weights:}
Instead of predicting object subgoals in $\mathrm{SE}(3)$, the high-level RL policy selects MPC weights $(w_{\text{pos}}, w_{\text{ori}})$ to implicitly define the subgoal: 
{\small
\begin{equation}
\label{equ.terminal_cost_repar}
V(\boldsymbol{q}^{\text{obj}},\boldsymbol{\bar{q}}^{\text{obj}}) {=} w_{\text{pos}}\,e_{\text{pos}}(\mathbf{p}^{\text{obj}},\mathbf{\bar p}^{\text{target}}) {+} w_{\text{ori}}\,e_{\text{ori}}(\boldsymbol{r}^{\text{obj}},\boldsymbol{\bar r}^{\text{target}}),
\end{equation}}

where $e_{\text{pos}}$ and $e_{\text{ori}}$ measure the object's position and orientation errors to the final target. For example,  $(w_{\text{pos}}, w_{\text{ori}})=(1,0)$ states the subgoal to be  target position-only. Thus, the high-level RL policy  predicts $(\boldsymbol{\bar{c}}^1, \boldsymbol{\bar{c}}^2, ..., \boldsymbol{\bar{c}}^n, w_{\text{pos}}, w_{\text{ori}})=\pi(\boldsymbol{o})$.

\emph{Option 2: Selecting from a pre-defined discrete subgoal candidate set:} 
For tasks, one can predefine a candidate subgoal set: 
$\mathcal{Q}=\{\boldsymbol{\bar{q}}^{\text{obj}}_1,\boldsymbol{\bar{q}}^{\text{obj}}_2,\ldots,\boldsymbol{\bar{q}}^{\text{obj}}_M\}$,
where each candidate corresponds to a meaningful intermediate object pose towards the final target. The high-level RL policy selects one candidate in $\mathcal{Q}$ to form the contact intention: $(\boldsymbol{\bar{c}}^1, \boldsymbol{\bar{c}}^2, \ldots, \boldsymbol{\bar{c}}^n, \, \boldsymbol{\bar{q}}^{\text{obj}}) = \pi(\boldsymbol{o}), \ \boldsymbol{\bar{q}}^{\text{obj}} \in \mathcal{Q}.$

\vspace{-8pt}
\section{Contact-Implicit Model Predictive Control}
\label{sec.MPC}
\vspace{-8pt}
We use ComFree-MPC for contact-implicit MPC, whose cost is conditioned on RL-predicted contact intentions.  ComFree-MPC performs high-speed  receding-horizon planning and control (around 100Hz), enabling the system to dynamically reason over contact modes (sliding, sticking,   breaking) and move the object to the subgoal. We details two key components of the ComFree MPC in (\ref{equ.mpc}).

\emph{(I) Complementarity-Free Contact Dynamics:} We adopt the quasi-dynamic complementarity-free contact model from \cite{jin2024complementarity} for our relatively slow non-prehensile manipulation task. The formulation of this contact dynamics model is shown in Appendix \ref{appendix:mpc_dyn}. 

\emph{(II) Cost function:} The {running cost} is  defined as $\sum\nolimits_{i=1}^n L(\mathbf{p}_t^{{\mathrm{ee}},i},\boldsymbol{\bar{c}}^i_k)= \sum\nolimits_{i=1}^n \, w_c\big\lVert \mathbf{p}_t^{{\mathrm{ee}},i} - T_{\text{obj}}^{w} \boldsymbol{\bar{c}}^i \big\rVert^2$, where  $T_{obj}^{w}$ transforms $\boldsymbol{\bar{c}}^i$ to the world frame. $w_c$ is the hyperparameter controlling the strength of this cost. This is to  encourage the robot end-effectors to stick to the contact position. The terminal cost is defined as (\ref{equ.terminal_cost_repar}) if RL action for subgoal is using Option 1. Otherwise if the RL action directly predicts the object subgoal  $\boldsymbol{\bar{q}}^{\text{obj}}$  (Option 2),  $\small V(\mathbf{q}_{H}^{\text{obj}}, \mathbf{\bar{q}}^{\text{obj}})$ directly penalizes the distance to   $\boldsymbol{\bar{q}}^{\text{obj}}$.
\vspace{-10pt}
\section{Experiment for Non-Prehensile Manipulation}
\vspace{-8pt}
\label{sec.experiment}
\begin{wrapfigure}{r}{0.55\textwidth}
    \centering
    \includegraphics[width=\linewidth]{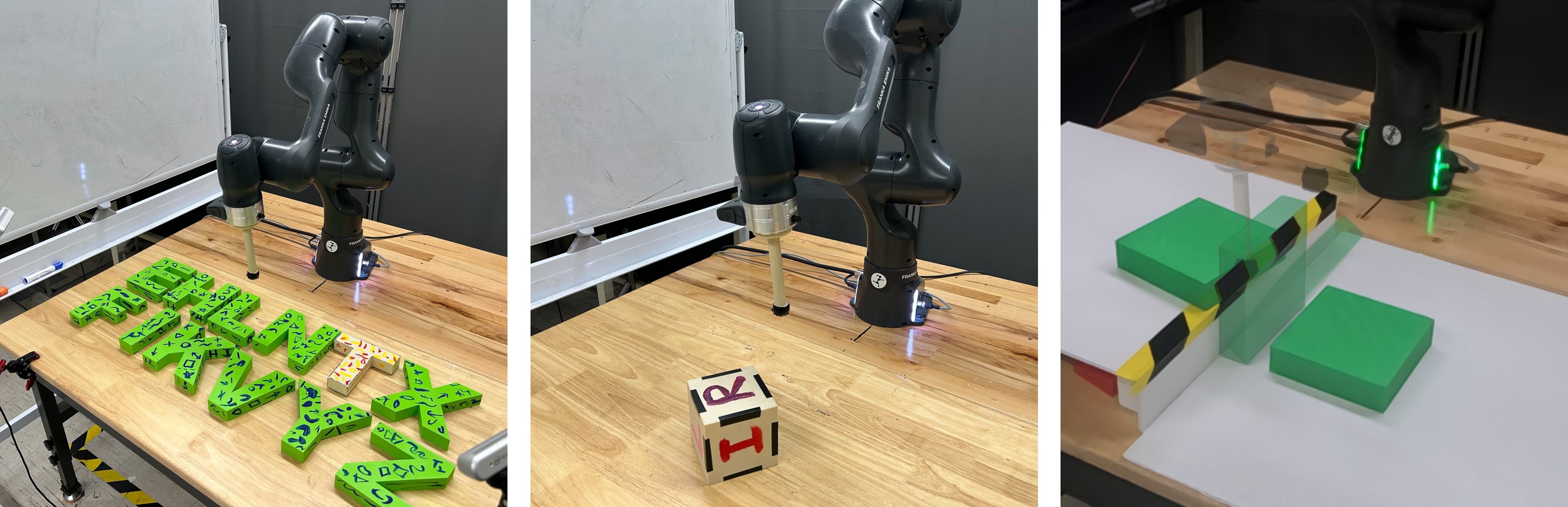}
    \caption{Left: geometry-generalized pushing. Mid: pivoting-flipping based object reorientation. Right: environment-assisted object repositioning.}
    \vspace{-15pt}
    \label{fig.arm_tasks}
\end{wrapfigure}

We focuses on 3 non-prehensile manipulation tasks (Fig.~\ref{fig.arm_tasks}). All  tasks use a 7-DOF Franka robot arm  with a stick end-effector.

\vspace{-5pt}
\paragraph{Geometry-generalized pushing}
The goal is to learn \emph{one policy} that pushes diverse letter-shaped objects from random initial poses to planar targets. We use 12  letters:  (E, H, L, N, T, X) for training and (F, I, K, V, Y, Z) held out for  generalization test. The task requires geometry-aware contact reasoning: sequentially selecting shape facets  and  proper contact  strategies. 

\vspace{-10pt}
\paragraph{Pivoting-flipping based object reorientation}
The goal is to reorient a cube from a random initial pose to a random 3D target. This task evaluates the method's ability to generate richer contact strategies  beyond planar pushing, e.g., toppling, pivoting, and sliding through robot-cube-table contact.
\vspace{-10pt}
\paragraph{Environment-assisted object repositioning}
The goal is to reorient a rectangular object from a random initial pose on an stair to a flip target pose on the table. This  requires  reasoning over environment geometry and exploit   environment--object contacts, e.g., by first pivoting the object by 90 degrees against the stair before completing the flip.


\vspace{-8pt}
\subsection{Training High-Level RL
Policy  in Abstract-Robot Simulation}
\vspace{-8pt}\label{sec:task_rl_training}
Due to the embodiment-agnostic nature of high-level contact intention prediction (except the number of active end-effector),  we train the high-level policy in simulation  where the robot end-effector is abstracted as a mass--spring--damper point  (Appendix \ref{appendix:task_setting}). This further reduces low-level MPC solve time, speeding up learning. 
We train the high-level policy using PPO \cite{schulman2017proximal}. At each episode, object and target poses are randomized (Appendix \ref{appendix:task_setting}), with one training object sampled per episode for the geometry-generalized pushing task. At the start of each episode, we perform domain randomization on mass, friction and end-effector actuation.
Details on rewards, hyperparameters, domain randomization and their implementation are in Appendix \ref{appendix:rl_setting}.

\begin{wrapfigure}{r}{0.7\textwidth}
\vspace{-10pt}
    \centering
    \begin{subfigure}[b]{0.32\linewidth}
    \includegraphics[width=\linewidth]{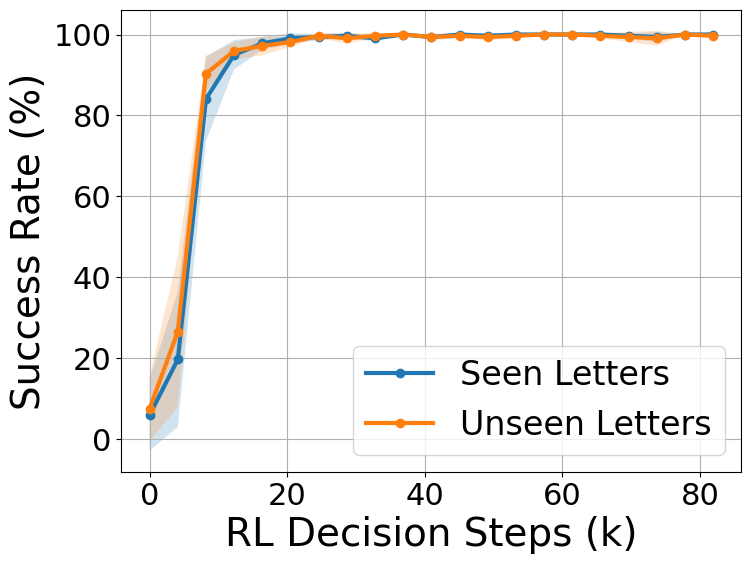}
  \end{subfigure}
  \begin{subfigure}[b]{0.32\linewidth}
    \includegraphics[width=\linewidth]{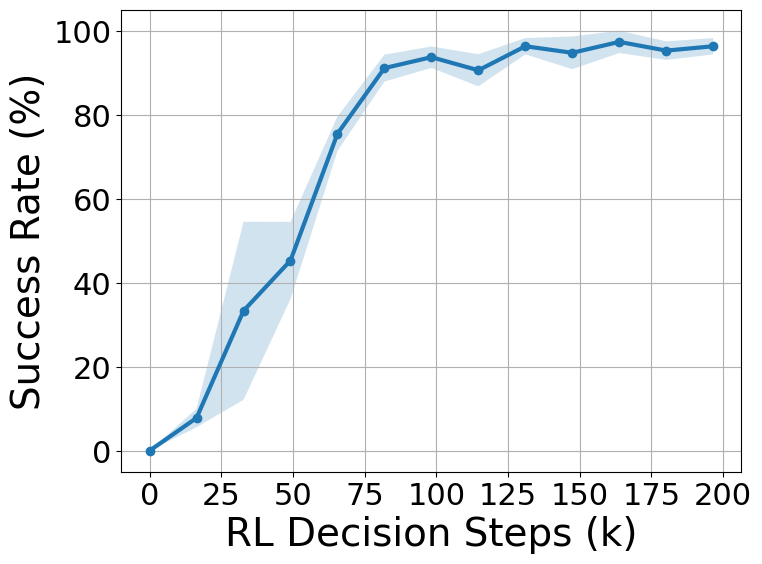}
  \end{subfigure}
  \begin{subfigure}[b]{0.32\linewidth}
    \includegraphics[width=\linewidth]{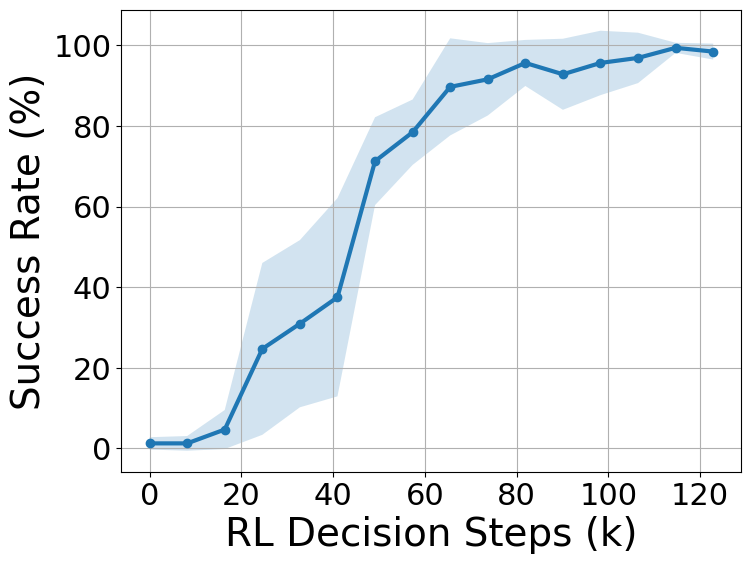}
  \end{subfigure}
    \caption{Learning curves for three tasks. Left: Geometry generalized pushing. Mid: Object reorientation. Right: env-assisted object repositioning.}
    \vspace{-10pt}
    \label{fig.curves}
\end{wrapfigure}

We perform 5 independent training runs per task, evaluating success rate at saved checkpoints with 64 trials each, and each trial uses random start/target poses.  In geometry-generalized pushing, we report separate curves for seen and unseen random letters. A trial succeeds if the object reaches sufficiently close to the target within 64 RL policy steps. Pose distributions and success criteria are in Appendix \ref{appendix:task_setting}, and learning curves with the abstract end-effector are shown in Fig.~\ref{fig.curves}.

The learning results consisitently show that our method achieves high success rates with  small amount of training data. The pushing task achieves $100\% {\pm} 0\%$ for seen and $99.69\% {\pm} 0.62\%$ for unseen letters after $\sim$40k; object 3D reorientation achieves $96.56\% {\pm} 2.50\%$ at $\sim$200k; and environment-assisted object repositioning task reaches $98.44\% {\pm} 1.98\%$ after $\sim$120k RL steps. 

\vspace{-8pt}
\subsection{Zero-Shot Transfer to Full Robotic Systems}
\vspace{-8pt}
\label{sec.sim_arm}
Deploying the  high-level policy trained in abstract-robot simulation to full robotic manipulation requires only adding the robot-specific ComFree-MPC without additional training. For one   step of the RL policy,  the low-level ComFree-MPC executes for multiple control steps (Appendix \ref{appendix:mpc_setting}). Task-wise ComFree-MPC parameters are set for different system (Appendix \ref{appendix:mpc_setting}).

\begin{wraptable}{r}{0.52\textwidth}
    \centering
    \vspace{-10pt}
    \caption{Result in Simulated Abstract/Full Setting}
    \tiny
    \begin{tabular}{llll}
    \toprule
    \textbf{Task} & \textbf{Success (Abstract)} & \textbf{Success (Full)} & \textbf{Steps (Full)}\\
    \midrule
    Pushing  (seen) & $100\% \pm 0\%$ & $100\% \pm 0\%$ & $7.4 \pm 6.9$\\
    Pushing (unseen) & $99.69\% \pm 0.62\%$ & $100\% \pm 0\%$ & $6.8 \pm 4.6$\\
    3D Reorientation & $96.56\% \pm 2.50\%$ & $95\% \pm 5.48\%$ & $14.5 \pm 9.5$\\
    Env-contact task & $98.44\% \pm 1.98\%$ & $95\% \pm 3.16\%$ & $4.0 \pm 2.0$\\
    \bottomrule
    \end{tabular}
    \vspace{-10pt}
    \label{tbl.abstrat_full_result}
\end{wraptable}
As shown in Fig. \ref{tbl.abstrat_full_result},  we  evaluate the trained RL policies in a full robot-arm simulation environment. We run 5 evaluation trials per letter in the geometry-generalized pushing task (abbrev. as Pushing), and 20  trials in both  {pivoting-flipping based reorientation task} (3D reorientation) and {environment-assisted object repositioning task} (Env-contact task). The learned policies achieve consistently  $\geq 95\%$ success rate, both in abstract setting and after zero-shot transfer to full robot setting, while requiring only a small number of high-level RL decision steps to complete each task (approximately 10 on average). Table~\ref{tbl.abstrat_full_result} summarizes the success rates of our method in both the abstract and full-system settings, along with the number of RL decision steps required for task completion in full-system setting.

\vspace{-8pt}
\subsection{Baseline Comparison}
\begin{wrapfigure}{r}{0.53\textwidth}
    \centering
    \begin{subfigure}[b]{0.48\linewidth}
    \includegraphics[width=\linewidth]{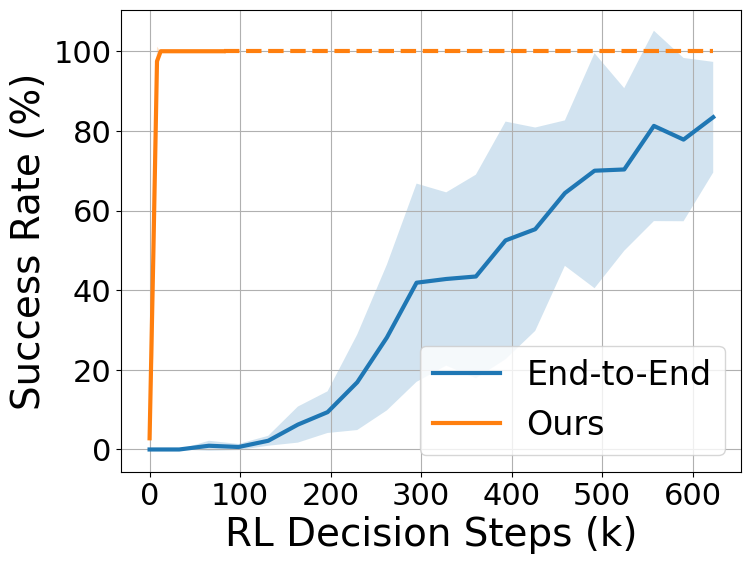}
  \end{subfigure}
  \begin{subfigure}[b]{0.48\linewidth}
    \includegraphics[width=\linewidth]{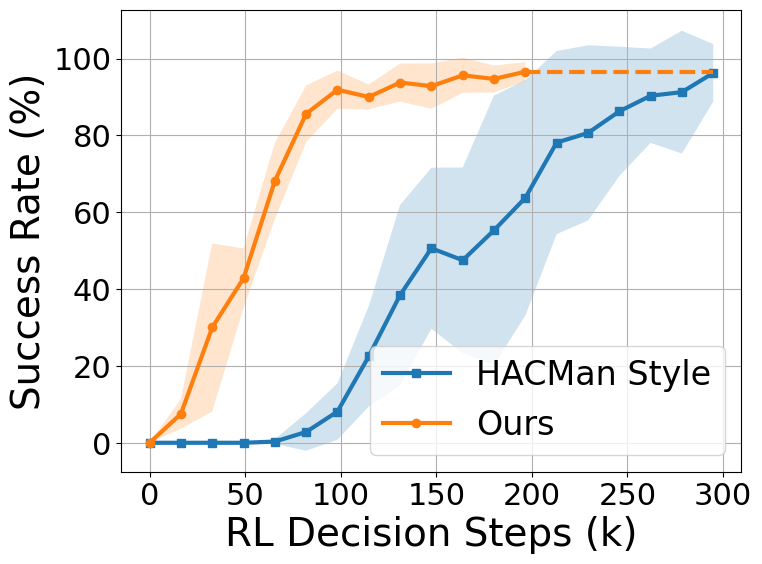}
  \end{subfigure}
    \caption{The result of baseline comparisons. Left:  comparison with end-to-end baseline, evaluated in the pushing T task. Right:  Comparison with HACMan-style  policy, evaluated in pivoting-flipping based object reorientation.}
    \vspace{-10pt}
    \label{fig.compare_end2end}
\end{wrapfigure}

\vspace{-8pt}
\label{sec:sim_baseline}
We compare our hierarchical RL-MPC method with baselines in abstract end-effector setting: (i) an  end-to-end RL policy, (ii) a HACMan-style  policy \cite{zhou2023hacman}, and (iii)  MPC-only baseline without high-level contact intention guidance. In each comparison, five independent training runs are performed, and performance is reported at checkpoints using 64 trials per run.
\vspace{-8pt}
\paragraph{Comparison with End-to-end RL learning}
The end-to-end policy directly predicts  end-effector actions, using an observation that includes object keypoints, goal flow, per-point clearance, and end-effector flow (distance vectors between the end-effector  and all keypoints) in  world frame. End effector flow provides spatial context about the robot–object geometry. See Appendix~\ref{appendix:end2end} for more detailed implementation.
Fig.~\ref{fig.compare_end2end} shows learning curves for  T-Pushing task  with abstract end-effector. 
The end-to-end policy converges slowly, reaching $83.44\% \pm 13.89\%$ after $\sim$600K RL steps (corresponding to 600K control steps), while our proposed method achieves $100\% \pm 0\%$ in 15K steps (approximately 300K control steps), giving $\sim$40$\times$ improvement in  terms of RL decision steps and $\sim$2$\times$ with respect to control steps.

\vspace{-8pt}
\paragraph{Comparison with HACMan-Style Method} We compare our RL-MPC framework with a HACMan-style  RL  policy~\cite{zhou2023hacman}, which predict both the contact locations and  end-effector  displacements. The RL observation space is the same as ours, while the action space is $\mathcal{P} \times [-0.1,0.1]^3$, where $\mathcal{P}$ denotes the set of candidate contact keypoints and \([-0.1,0.1]^3\) is the range of end-effector displacement. 
The learning curve for solving the pivoting-flipping object reorientation  with abstract end-effector is  in Fig. \ref{fig.compare_end2end} (Right).
HACMan-style hierarchical policy requires roughly 300K RL steps to reach about $96.25\% \pm 7.50 \%$ success rate, while our method converges to $96.56\% \pm 2.50\%$ success rate within 200K steps, showing significantly faster convergence speed. Furthermore, performance variance of our methods is shown  considerably  small. This is mainly because the high-variance contact dynamics reasoning is reliably handled by our MPC level, instead of learned via RL in HACMan policy.

\vspace{-8pt}
\paragraph{Compared with MPC-only baseline}
We also compare with MPC-only baseline that directly control toward the final object target without high-level contact-intention guidance. We evaluate this  on the simulated letter-T pushing task using the abstract end-effector setting. In each episode, the end effector is randomly initialized near the object. The MPC-only baseline almost always fails to complete the task, achieving zero success rate. This is because of the short-sighted nature of contact-implicit MPC planning, which necessitates the imprtance of high-level geometry planning.

\vspace{-5pt}
\subsection{Ablation Studies}
\begin{wrapfigure}{r}{0.55\textwidth}
    \centering
    \begin{subfigure}[b]{0.48\linewidth}
    \includegraphics[width=\linewidth]{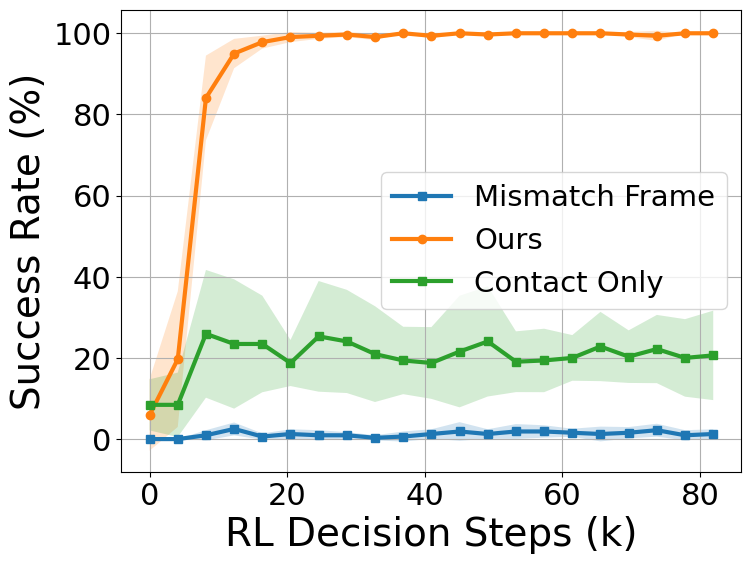}
  \end{subfigure}
  \begin{subfigure}[b]{0.48\linewidth}
    \includegraphics[width=\linewidth]{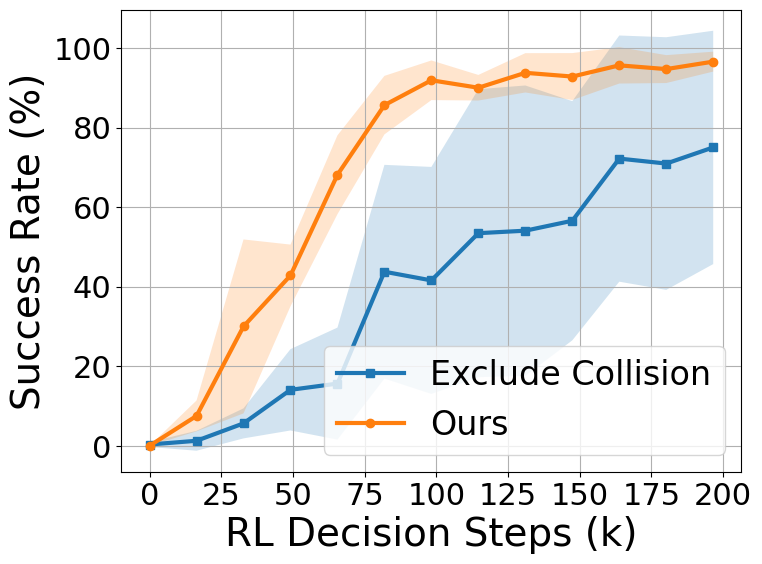}
  \end{subfigure}
    \caption{The result of ablation studies. Left: Ablation 1 \& 2 in geometry-generalized pushing task. Right: Ablation 3 in pivoting-flipping based object reorientation.}
    \vspace{-10pt}
    \label{fig.ablation}
\end{wrapfigure}

\vspace{-8pt}
\paragraph{Ablation 1: Removing Object Subgoal in the Contact Intention}
We consider a variant where the high-level policy predicts  contact locations only. In this case, the low-level MPC always moves the object to the final target. We evaluate this ablation using the  simulated geometry-generalized pushing task with the abstract end-effector (the same setting as the previous experiment). As shown by the green curve in Fig.~\ref{fig.ablation} (left, ``contact-only" curve), removing subgoal prediction drops the success rate from $100\% \pm 0.0\%$  to $20.62\% \pm 11.02\%$. We find that in the contact-location-only case, the whole hierarchical policy often leads to inefficient pushing (usually exceeding the 64-step epsiode termination threshold). This is because the policy only specifies where to contact, while the MPC must directly optimize toward the final pose, potentially leading the object pose to  local minima that have little contribution to the overall task progress. The ablation result highlights the critical role of subgoal selection for efficient  manipulation.
\vspace{-10pt}
\paragraph{Ablation 2: World-Frame Observation}
This ablation  is designed to highlight the importance of expressing the geometry component $\mathcal{P}$ and the target component $\Delta \mathcal{P}$ in the object frame. We train a policy under the same settings as the main experiment for the {simulated geometry-generalized pushing} task with abstract end-effector, with the only difference being that $\Delta \mathcal{P}$ is expressed in the world while $\mathcal{P}$ remains  in the object frame. As shown with the blue curve in Fig. \ref{fig.ablation}(left, ``Mismatch frame" curve), learning fails. This result demonstrates that a consistent object-centric representation is crucial for stable learning and for enabling the policy to reason effectively about geometry without confusing with the object-pose-dependent data.

\vspace{-10pt}
\paragraph{Ablation 3: Removing Collision Component in Observation}
To evaluate the role of the \emph{collision component} $\mathcal{D}_k$, we remove it from the observation for {simulated pivoting-flipping object reorientation} task with abstract end-effector, keeping all other settings unchanged. As shown in Fig.~\ref{fig.ablation}(right, ``Exclude Collision" curves), both learning speed and success rate drop (the success rate drops from $96.56\% \pm 2.50\%$   to  $75.00\% \pm 29.33\%$). A possible explanation is that, without the collision component, the RL policy struggles to identify feasible contact locations and effectively exploit object--environment contacts.

\vspace{-10pt}
\section{Zero-Shot Sim-To-Real Transfer to Hardware}
\vspace{-10pt}
Across all three tasks, we directly transfer the high-level RL policy trained in abstract-robot simulation  to the real-world hardware. The transfer is zero-shot without any fine-tuning/post-training. It only requires to add a ComFree-MPC controller with few physical parameters minimally tuned for the hardware (e.g., few mass and stiffness parameters). See  details in Appendix \ref{appendix:mpc_setting}.

\begin{wraptable}{r}{0.35\textwidth}
\centering
\vspace{-10pt}
\caption{Result in real-world non-prehensile manipulation}
\label{tbl.real_result}
\tiny
\begin{tabular}{lcc}
\toprule
\textbf{Task} & \textbf{Success rate} & \textbf{Num. steps}\\
\midrule
Pushing (seen) & 100.0\% & 6.10 $\pm$ 3.40 \\
Pushing (unseen) & 95.0\% & 7.35 $\pm$ 4.73 \\
3D Reorientation & 100.0\% & 24.96 $\pm$ 15.5 \\
Env-contact task & 88.0\% & 5.92 $\pm$ 4.35 \\
\bottomrule
\end{tabular}
\vspace{-15pt}
\end{wraptable}

On hardware evaluation, we perform 10 independent trials for each letter in pushing task, 25 trials for both pivoting-flipping based object reorientation task and environment-assisted object repositioning. Other evaluation protocols follow those in simulation.   Table \ref{tbl.real_result} shows the hardware results. It reports with two metrics: success rate over a fixed number of trials, and the number of RL decision steps needed to complete the task (num\_steps). Our method shows robust zero-shot sim-to-real transfer  with high success rates. In the pushing tasks, most objects, including unseen ones, achieve 100\% success with few steps, showing reliable sim-to-real transfer; only letter I  occasionally fails due to  tracking failure. In the object reorientation task, our method maintains strong success, though requiring more steps. In the environment-assisted repositioning, the real-world success rate is slight lower to $88\%$, which is likely caused by vision tracking error.


\vspace{-10pt}
\paragraph{Qualitative: diverse contact interaction emergence} In the RL-MPC real-world deployment,  rich contact strategies emerge for different tasks.  
In the object reorientation task,  shown in Fig.~\ref{fig.flip_modes} of Appendix \ref{appendix:realworld_fig}, the robot end effector reorientates the cube purely by frictional sliding on the top face (first row in Fig.~\ref{fig.flip_modes}), pivots the cube at an edge to rotate the object (second row), or pivots the cube at one corner (third row).  In the environment-assisted repositioning task, the robot  leverages the stair edge and table surface as  source of contact constraints, sliding, pivoting, nudging, and toppling  the object  to achieve the flip task. An example can be found in Fig. \ref{fig.edge_real} of Appendix. \ref{appendix:realworld_fig}. 

\vspace{-10pt}
\paragraph{Baseline Comparison} Similar to in simulation, we also compare our method with the end-to-end and MPC-only baselines in real-world experiments. Both  are tested in pushing-T task. The end-to-end baseline achieves a $35\%$ success rate, while the MPC-only baseline achieves $0\%$. This again shows advantages of the proposed methods. Details are in the Appendix \ref{appendix:realworld_baseline}.

\vspace{-8pt}
\section{Conclusion and Limitations}
\vspace{-8pt}
We introduced a hierarchical RL-MPC framework for contact-rich manipulation that explicitly separates geometry reasoning  from  contact dynamics planning and control. The high-level RL policy predicts a \emph{contact intention}, including an object-surface contact location and a post-contact object-level subgoal, while a low-level contact-implicit MPC  performs online contact dynamics planning  to  realize each intention. Across non-prehensile manipulation tasks, our approach achieves high success rate with substantially less training data, strong robustness, and zero-shot sim-to-real transfer. 

The proposed approach  has two potential limitations. First, it relies on accurate object pose estimation for computing goal flow and MPC terms, whose performance can be affected by tracking stability. Second, using a discrete keypoint set $\mathcal{P}$ simplifies learning but could scales poorly with multiple end-effectors due to the combinatorial action space. Future work will address those issues and extend the framework for  multi-fingered dexterous manipulation.

\bibliography{references}  

@article{grafton2010cognitive,
  title={The cognitive neuroscience of prehension: recent developments},
  author={Grafton, Scott T},
  journal={Experimental brain research},
  volume={204},
  number={4},
  pages={475--491},
  year={2010},
  publisher={Springer}
}

@article{merel2019hierarchical,
  title={Hierarchical motor control in mammals and machines},
  author={Merel, Josh and Botvinick, Matthew and Wayne, Greg},
  journal={Nature communications},
  volume={10},
  number={1},
  pages={5489},
  year={2019},
  publisher={Nature Publishing Group UK London}
}

@article{jin2024complementarity,
  title={Complementarity-free multi-contact modeling and optimization for dexterous manipulation},
  author={Jin, Wanxin},
  journal={arXiv preprint arXiv:2408.07855},
  year={2024}
}

@article{zhou2023hacman,
  title={Hacman: Learning hybrid actor-critic maps for 6d non-prehensile manipulation},
  author={Zhou, Wenxuan and Jiang, Bowen and Yang, Fan and Paxton, Chris and Held, David},
  journal={arXiv preprint arXiv:2305.03942},
  year={2023}
}

@software{Wenhao_Yan_and_Jie_Chu_FoundationPose_2025,
author = {Wenhao Yan and Jie Chu},
license = {MIT},
month = mar,
title = {{FoundationPose++}},
url = {https://github.com/teal024/FoundationPose-plus-plus},
year = {2025}
}

@inproceedings{huang2010lcm,
  title={LCM: Lightweight communications and marshalling},
  author={Huang, Albert S and Olson, Edwin and Moore, David C},
  booktitle={2010 IEEE/RSJ International Conference on Intelligent Robots and Systems},
  pages={4057--4062},
  year={2010},
  organization={IEEE}
}

@article{schulman2017proximal,
  title={Proximal policy optimization algorithms},
  author={Schulman, John and Wolski, Filip and Dhariwal, Prafulla and Radford, Alec and Klimov, Oleg},
  journal={arXiv preprint arXiv:1707.06347},
  year={2017}
}

@article{wachter2006implementation,
  title={On the implementation of an interior-point filter line-search algorithm for large-scale nonlinear programming},
  author={W{\"a}chter, Andreas and Biegler, Lorenz T},
  journal={Mathematical programming},
  volume={106},
  number={1},
  pages={25--57},
  year={2006},
  publisher={Springer}
}

@Article{CasADi,
  author = {Joel A E Andersson and Joris Gillis and Greg Horn
            and James B Rawlings and Moritz Diehl},
  title = {{CasADi} -- {A} software framework for nonlinear optimization
           and optimal control},
  journal = {Mathematical Programming Computation},
  volume = {11},
  number = {1},
  pages = {1--36},
  year = {2019},
  publisher = {Springer},
  doi = {10.1007/s12532-018-0139-4}
}

@article{yang2025twintrack,
  title={TwinTrack: Bridging Vision and Contact Physics for Real-Time Tracking of Unknown Dynamic Objects},
  author={Yang, Wen and Xie, Zhixian and Zhang, Xuechao and Amor, Heni Ben and Lin, Shan and Jin, Wanxin},
  journal={arXiv preprint arXiv:2505.22882},
  year={2025}
}

@article{elsner2023taming,
title = {Taming the Panda with Python: A powerful duo for seamless robotics programming and integration},
journal = {SoftwareX},
volume = {24},
pages = {101532},
year = {2023},
issn = {2352-7110},
doi = {https://doi.org/10.1016/j.softx.2023.101532},
url = {https://www.sciencedirect.com/science/article/pii/S2352711023002285},
author = {Jean Elsner}
}

@article{stable-baselines3,
  author  = {Antonin Raffin and Ashley Hill and Adam Gleave and Anssi Kanervisto and Maximilian Ernestus and Noah Dormann},
  title   = {Stable-Baselines3: Reliable Reinforcement Learning Implementations},
  journal = {Journal of Machine Learning Research},
  year    = {2021},
  volume  = {22},
  number  = {268},
  pages   = {1-8},
  url     = {http://jmlr.org/papers/v22/20-1364.html}
}

@article{zhao2023learning,
  title={Learning fine-grained bimanual manipulation with low-cost hardware},
  author={Zhao, Tony Z and Kumar, Vikash and Levine, Sergey and Finn, Chelsea},
  journal={arXiv preprint arXiv:2304.13705},
  year={2023}
}

@article{chi2025diffusion,
  title={Diffusion policy: Visuomotor policy learning via action diffusion},
  author={Chi, Cheng and Xu, Zhenjia and Feng, Siyuan and Cousineau, Eric and Du, Yilun and Burchfiel, Benjamin and Tedrake, Russ and Song, Shuran},
  journal={The International Journal of Robotics Research},
  volume={44},
  number={10-11},
  pages={1684--1704},
  year={2025},
  publisher={Sage Publications Sage UK: London, England}
}

@article{black2410pi0,
  title={$\pi$0: A vision-language-action flow model for general robot control. CoRR, abs/2410.24164, 2024. doi: 10.48550},
  author={Black, Kevin and Brown, Noah and Driess, Danny and Esmail, Adnan and Equi, Michael and Finn, Chelsea and Fusai, Niccolo and Groom, Lachy and Hausman, Karol and Ichter, Brian and others},
  journal={arXiv preprint ARXIV.2410.24164}
}

@article{kim2024openvla,
  title={Openvla: An open-source vision-language-action model},
  author={Kim, Moo Jin and Pertsch, Karl and Karamcheti, Siddharth and Xiao, Ted and Balakrishna, Ashwin and Nair, Suraj and Rafailov, Rafael and Foster, Ethan and Lam, Grace and Sanketi, Pannag and others},
  journal={arXiv preprint arXiv:2406.09246},
  year={2024}
}

@article{figure2024helix,
  title={Helix: A vision-language-action model for generalist humanoid control},
  author={Figure, AI},
  journal={Figure AI News},
  year={2024}
}

@inproceedings{qi2023hand,
  title={In-hand object rotation via rapid motor adaptation},
  author={Qi, Haozhi and Kumar, Ashish and Calandra, Roberto and Ma, Yi and Malik, Jitendra},
  booktitle={Conference on Robot Learning},
  pages={1722--1732},
  year={2023},
  organization={PMLR}
}

@article{yin2023rotating,
  title={Rotating without seeing: Towards in-hand dexterity through touch},
  author={Yin, Zhao-Heng and Huang, Binghao and Qin, Yuzhe and Chen, Qifeng and Wang, Xiaolong},
  journal={arXiv preprint arXiv:2303.10880},
  year={2023}
}

@article{mason1986mechanics,
  title={Mechanics and planning of manipulator pushing operations},
  author={Mason, Matthew T},
  journal={The International Journal of Robotics Research},
  volume={5},
  number={3},
  pages={53--71},
  year={1986},
  publisher={Sage Publications Sage CA: Thousand Oaks, CA}
}

@article{lynch1996stable,
  title={Stable pushing: Mechanics, controllability, and planning},
  author={Lynch, Kevin M and Mason, Matthew T},
  journal={The international journal of robotics research},
  volume={15},
  number={6},
  pages={533--556},
  year={1996},
  publisher={Sage Publications Sage CA: Thousand Oaks, CA}
}

@article{akella1998posing,
  title={Posing polygonal objects in the plane by pushing},
  author={Akella, Srinivas and Mason, Matthew T},
  journal={The International Journal of Robotics Research},
  volume={17},
  number={1},
  pages={70--88},
  year={1998},
  publisher={Sage Publications Sage CA: Thousand Oaks, CA}
}

@inproceedings{mordatch2012contact,
  title={Contact-invariant optimization for hand manipulation},
  author={Mordatch, Igor and Popovi{\'c}, Zoran and Todorov, Emanuel},
  booktitle={Proceedings of the ACM SIGGRAPH/Eurographics symposium on computer animation},
  pages={137--144},
  year={2012}
}

@inproceedings{moura2022non,
  title={Non-prehensile planar manipulation via trajectory optimization with complementarity constraints},
  author={Moura, Joao and Stouraitis, Theodoros and Vijayakumar, Sethu},
  booktitle={2022 International Conference on Robotics and Automation (ICRA)},
  pages={970--976},
  year={2022},
  organization={IEEE}
}

@article{yi2023precise,
  title={Precise object sliding with top contact via asymmetric dual limit surfaces},
  author={Yi, Xili and Fazeli, Nima},
  journal={arXiv preprint arXiv:2305.14289},
  year={2023}
}

@inproceedings{wang2022contact,
  title={Contact-implicit planning and control for non-prehensile manipulation using state-triggered constraints},
  author={Wang, Maozhen and {\"O}nol, Aykut {\"O}zg{\"u}n and Long, Philip and Pad{\i}r, Ta{\c{s}}k{\i}n},
  booktitle={The International Symposium of Robotics Research},
  pages={189--204},
  year={2022},
  organization={Springer}
}

@article{posa2014direct,
  title={A direct method for trajectory optimization of rigid bodies through contact},
  author={Posa, Michael and Cantu, Cecilia and Tedrake, Russ},
  journal={The International Journal of Robotics Research},
  volume={33},
  number={1},
  pages={69--81},
  year={2014},
  publisher={Sage Publications Sage UK: London, England}
}

@article{yang2024dynamic,
  title={Dynamic on-palm manipulation via controlled sliding},
  author={Yang, William and Posa, Michael},
  year={2024},
  publisher={Robotics: Science and Systems Foundation}
}

@article{aydinoglu2024consensus,
  title={Consensus complementarity control for multi-contact mpc},
  author={Aydinoglu, Alp and Wei, Adam and Huang, Wei-Cheng and Posa, Michael},
  journal={IEEE Transactions on Robotics},
  year={2024},
  publisher={IEEE}
}

@article{bui2025push,
  title={Push Anything: Single-and Multi-Object Pushing From First Sight with Contact-Implicit MPC},
  author={Bui, Hien and Gao, Yufeiyang and Yang, Haoran and Cui, Eric and Mody, Siddhant and Acosta, Brian and Felix, Thomas Stephen and Bianchini, Bibit and Posa, Michael},
  journal={arXiv preprint arXiv:2510.19974},
  year={2025}
}

@article{kurtz2023inverse,
  title={Inverse dynamics trajectory optimization for contact-implicit model predictive control},
  author={Kurtz, Vince and Castro, Alejandro and {\"O}nol, Aykut {\"O}zg{\"u}n and Lin, Hai},
  journal={The International Journal of Robotics Research},
  pages={02783649251344635},
  year={2023},
  publisher={SAGE Publications Sage UK: London, England}
}

@article{cheng2023enhancing,
  title={Enhancing dexterity in robotic manipulation via hierarchical contact exploration},
  author={Cheng, Xianyi and Patil, Sarvesh and Temel, Zeynep and Kroemer, Oliver and Mason, Matthew T},
  journal={IEEE Robotics and Automation Letters},
  volume={9},
  number={1},
  pages={390--397},
  year={2023},
  publisher={IEEE}
}

@article{cho2025hierarchical,
  title={Hierarchical and Modular Network on Non-prehensile Manipulation in General Environments},
  author={Cho, Yoonyoung and Han, Junhyek and Han, Jisu and Kim, Beomjoon},
  journal={arXiv preprint arXiv:2502.20843},
  year={2025}
}

@article{li2025pin,
  title={Pin-wm: Learning physics-informed world models for non-prehensile manipulation},
  author={Li, Wenxuan and Zhao, Hang and Yu, Zhiyuan and Du, Yu and Zou, Qin and Hu, Ruizhen and Xu, Kai},
  journal={arXiv preprint arXiv:2504.16693},
  year={2025}
}

@article{cho2024corn,
  title={Corn: Contact-based object representation for nonprehensile manipulation of general unseen objects},
  author={Cho, Yoonyoung and Han, Junhyek and Cho, Yoontae and Kim, Beomjoon},
  journal={arXiv preprint arXiv:2403.10760},
  year={2024}
}

@article{lin2025sim,
  title={Sim-to-real reinforcement learning for vision-based dexterous manipulation on humanoids},
  author={Lin, Toru and Sachdev, Kartik and Fan, Linxi and Malik, Jitendra and Zhu, Yuke},
  journal={arXiv preprint arXiv:2502.20396},
  year={2025}
}

@article{saigusa2022imitation,
  title={Imitation learning for nonprehensile manipulation through self-supervised learning considering motion speed},
  author={Saigusa, Yuki and Sakaino, Sho and Tsuji, Toshiaki},
  journal={IEEE Access},
  volume={10},
  pages={68291--68306},
  year={2022},
  publisher={IEEE}
}

@inproceedings{zhu2023viola,
  title={Viola: Imitation learning for vision-based manipulation with object proposal priors},
  author={Zhu, Yifeng and Joshi, Abhishek and Stone, Peter and Zhu, Yuke},
  booktitle={Conference on Robot Learning},
  pages={1199--1210},
  year={2023},
  organization={PMLR}
}

@inproceedings{rouxel2024flow,
  title={Flow matching imitation learning for multi-support manipulation},
  author={Rouxel, Quentin and Ferrari, Andrea and Ivaldi, Serena and Mouret, Jean-Baptiste},
  booktitle={2024 IEEE-RAS 23rd International Conference on Humanoid Robots (Humanoids)},
  pages={528--535},
  year={2024},
  organization={IEEE}
}

@inproceedings{tosun2019pixels,
  title={Pixels to plans: Learning non-prehensile manipulation by imitating a planner},
  author={Tosun, Tarik and Mitchell, Eric and Eisner, Ben and Huh, Jinwook and Lee, Bhoram and Lee, Daewon and Isler, Volkan and Seung, H Sebastian and Lee, Daniel},
  booktitle={2019 IEEE/RSJ International Conference on Intelligent Robots and Systems (IROS)},
  pages={7431--7438},
  year={2019},
  organization={IEEE}
}

@article{sun2022integrating,
  title={Integrating reinforcement learning and learning from demonstrations to learn nonprehensile manipulation},
  author={Sun, Xilong and Li, Jiqing and Kovalenko, Anna Vladimirovna and Feng, Wei and Ou, Yongsheng},
  journal={IEEE Transactions on Automation Science and Engineering},
  volume={20},
  number={3},
  pages={1735--1744},
  year={2022},
  publisher={IEEE}
}

@inproceedings{yang2022fast,
  title={Fast and efficient locomotion via learned gait transitions},
  author={Yang, Yuxiang and Zhang, Tingnan and Coumans, Erwin and Tan, Jie and Boots, Byron},
  booktitle={Conference on robot learning},
  pages={773--783},
  year={2022},
  organization={PMLR}
}

@article{zhang2022model,
  title={Model predictive control of quadruped robot based on reinforcement learning},
  author={Zhang, Zhitong and Chang, Xu and Ma, Hongxu and An, Honglei and Lang, Lin},
  journal={Applied Sciences},
  volume={13},
  number={1},
  pages={154},
  year={2022},
  publisher={MDPI}
}

@inproceedings{chen2024learning,
  title={Learning agile locomotion and adaptive behaviors via rl-augmented mpc},
  author={Chen, Yiyu and Nguyen, Quan},
  booktitle={2024 IEEE International Conference on Robotics and Automation (ICRA)},
  pages={11436--11442},
  year={2024},
  organization={IEEE}
}

@techreport{jia2024rl,
  title={Rl-mpc: reinforcement learning aided model predictive controller for autonomous vehicle lateral control},
  author={Jia, Muye and Tao, Mingyuan and Xu, Meng and Zhang, Peng and Qiu, Jiayi and Bergsieker, Gerald and Chen, Jun},
  year={2024},
  institution={SAE Technical Paper}
}

@article{hu2025lno,
  title={LNO-Driven Deep RL-MPC: Hierarchical Adaptive Control Architecture for Dynamic Legged Locomotion},
  author={Hu, Lei and Ding, Liang and Yang, Huaiguang and Liu, Tie and Zhang, Ao and Chen, Siyang and Gao, Haibo and Xu, Peng and Deng, Zongquan},
  journal={IEEE Transactions on Industrial Informatics},
  year={2025},
  publisher={IEEE}
}

@inproceedings{liu2025autonomous,
  title={Autonomous Bimanual Manipulation of Deformable Objects Using Deep Reinforcement Learning Guided Adaptive Control},
  author={Liu, Jiayi and Yang, Sihang and Wang, Yiwei and Zhao, Huan and Ding, Han},
  booktitle={2025 IEEE International Conference on Robotics and Automation (ICRA)},
  pages={6904--6910},
  year={2025},
  organization={IEEE}
}

@inproceedings{shin2019autonomous,
  title={Autonomous tissue manipulation via surgical robot using learning based model predictive control},
  author={Shin, Changyeob and Ferguson, Peter Walker and Pedram, Sahba Aghajani and Ma, Ji and Dutson, Erik P and Rosen, Jacob},
  booktitle={2019 International conference on robotics and automation (ICRA)},
  pages={3875--3881},
  year={2019},
  organization={IEEE}
}

@inproceedings{omer2021model,
  title={Model predictive-actor critic reinforcement learning for dexterous manipulation},
  author={Omer, Muhammad and Ahmed, Rami and Rosman, Benjamin and Babikir, Sharief F},
  booktitle={2020 International Conference on Computer, Control, Electrical, and Electronics Engineering (ICCCEEE)},
  pages={1--6},
  year={2021},
  organization={IEEE}
}

@article{lopes2025model,
  title={Model-Based Lookahead Reinforcement Learning for in-hand manipulation},
  author={Lopes, Alexandre and Barata, Catarina and Moreno, Plinio},
  journal={arXiv preprint arXiv:2510.08884},
  year={2025}
}

@article{bing2023safety,
  title={Safety guaranteed manipulation based on reinforcement learning planner and model predictive control actor},
  author={Bing, Zhenshan and Mavrichev, Aleksandr and Shen, Sicong and Yao, Xiangtong and Chen, Kejia and Huang, Kai and Knoll, Alois},
  journal={arXiv preprint arXiv:2304.09119},
  year={2023}
}

@inproceedings{zhuang2025rm,
  title={RM-Planner: Integrating Reinforcement Learning with Whole-Body Model Predictive Control for Mobile Manipulation},
  author={Zhuang, Zixuan and Zheng, Le and Li, Wanyue and Liu, Renming and Lu, Peng and Cheng, Hui},
  booktitle={2025 IEEE International Conference on Robotics and Automation (ICRA)},
  pages={7263--7269},
  year={2025},
  organization={IEEE}
}

@inproceedings{hermans2011affordance,
  title={Affordance prediction via learned object attributes},
  author={Hermans, Tucker and Rehg, James M and Bobick, Aaron},
  booktitle={IEEE international conference on robotics and automation (ICRA): Workshop on semantic perception, mapping, and exploration},
  pages={181--184},
  year={2011},
  organization={Citeseer}
}

@article{geng2022end,
  title={End-to-end affordance learning for robotic manipulation},
  author={Geng, Yiran and An, Boshi and Geng, Haoran and Chen, Yuanpei and Yang, Yaodong and Dong, Hao},
  journal={arXiv preprint arXiv:2209.12941},
  year={2022}
}

@article{xu2021affordance,
  title={An affordance keypoint detection network for robot manipulation},
  author={Xu, Ruinian and Chu, Fu-Jen and Tang, Chao and Liu, Weiyu and Vela, Patricio A},
  journal={IEEE Robotics and Automation Letters},
  volume={6},
  number={2},
  pages={2870--2877},
  year={2021},
  publisher={IEEE}
}

@article{chu2019toward,
  title={Toward affordance detection and ranking on novel objects for real-world robotic manipulation},
  author={Chu, Fu-Jen and Xu, Ruinian and Seguin, Landan and Vela, Patricio A},
  journal={IEEE Robotics and Automation Letters},
  volume={4},
  number={4},
  pages={4070--4077},
  year={2019},
  publisher={IEEE}
}

@article{chu2019learning,
  title={Learning affordance segmentation for real-world robotic manipulation via synthetic images},
  author={Chu, Fu-Jen and Xu, Ruinian and Vela, Patricio A},
  journal={IEEE Robotics and Automation Letters},
  volume={4},
  number={2},
  pages={1140--1147},
  year={2019},
  publisher={IEEE}
}

@article{jiang2021synergies,
  title={Synergies between affordance and geometry: 6-dof grasp detection via implicit representations},
  author={Jiang, Zhenyu and Zhu, Yifeng and Svetlik, Maxwell and Fang, Kuan and Zhu, Yuke},
  journal={arXiv preprint arXiv:2104.01542},
  year={2021}
}

@inproceedings{li2025learning,
  title={Learning precise affordances from egocentric videos for robotic manipulation},
  author={Li, Gen and Tsagkas, Nikolaos and Song, Jifei and Mon-Williams, Ruaridh and Vijayakumar, Sethu and Shao, Kun and Sevilla-Lara, Laura},
  booktitle={Proceedings of the IEEE/CVF International Conference on Computer Vision},
  pages={10581--10591},
  year={2025}
}

@article{huang2023grounded,
  title={Grounded decoding: Guiding text generation with grounded models for embodied agents},
  author={Huang, Wenlong and Xia, Fei and Shah, Dhruv and Driess, Danny and Zeng, Andy and Lu, Yao and Florence, Pete and Mordatch, Igor and Levine, Sergey and Hausman, Karol and others},
  journal={Advances in Neural Information Processing Systems},
  volume={36},
  pages={59636--59661},
  year={2023}
}

@inproceedings{tang2025roboafford,
  title={Roboafford: A dataset and benchmark for enhancing object and spatial affordance learning in robot manipulation},
  author={Tang, Yingbo and Zhang, Lingfeng and Zhang, Shuyi and Zhao, Yinuo and Hao, Xiaoshuai},
  booktitle={Proceedings of the 33rd ACM International Conference on Multimedia},
  pages={12706--12713},
  year={2025}
}

@article{song2025maniplvm,
  title={Maniplvm-r1: Reinforcement learning for reasoning in embodied manipulation with large vision-language models},
  author={Song, Zirui and Ouyang, Guangxian and Li, Mingzhe and Ji, Yuheng and Wang, Chenxi and Xu, Zixiang and Zhang, Zeyu and Zhang, Xiaoqing and Jiang, Qian and Chen, Zhenhao and others},
  journal={arXiv preprint arXiv:2505.16517},
  year={2025}
}

@inproceedings{ji2025robobrain,
  title={Robobrain: A unified brain model for robotic manipulation from abstract to concrete},
  author={Ji, Yuheng and Tan, Huajie and Shi, Jiayu and Hao, Xiaoshuai and Zhang, Yuan and Zhang, Hengyuan and Wang, Pengwei and Zhao, Mengdi and Mu, Yao and An, Pengju and others},
  booktitle={Proceedings of the Computer Vision and Pattern Recognition Conference},
  pages={1724--1734},
  year={2025}
}

@article{tong2024oval,
  title={Oval-prompt: Open-vocabulary affordance localization for robot manipulation through llm affordance-grounding},
  author={Tong, Edmond and Opipari, Anthony and Lewis, Stanley and Zeng, Zhen and Jenkins, Odest Chadwicke},
  journal={arXiv preprint arXiv:2404.11000},
  year={2024}
}

@inproceedings{li2024manipllm,
  title={Manipllm: Embodied multimodal large language model for object-centric robotic manipulation},
  author={Li, Xiaoqi and Zhang, Mingxu and Geng, Yiran and Geng, Haoran and Long, Yuxing and Shen, Yan and Zhang, Renrui and Liu, Jiaming and Dong, Hao},
  booktitle={Proceedings of the IEEE/CVF Conference on Computer Vision and Pattern Recognition},
  pages={18061--18070},
  year={2024}
}

@inproceedings{wen2024foundationpose,
  title={Foundationpose: Unified 6d pose estimation and tracking of novel objects},
  author={Wen, Bowen and Yang, Wei and Kautz, Jan and Birchfield, Stan},
  booktitle={Proceedings of the IEEE/CVF Conference on Computer Vision and Pattern Recognition},
  pages={17868--17879},
  year={2024}
}

@inproceedings{pyg,
  title={{PyG} 2.0: Scalable Learning on Real World Graphs},
  author={Fey, Matthias and Sunil, Jinu and Nitta, Akihiro and Puri, Rishi and Shah, Manan and Stojanovi\v{c}, Bla\v{z} and Bendias, Ramona and Alexandria, Barghi and Kocijan, Vid and Zhang, Zecheng and He, Xinwei and Lenssen, Jan E. and Leskovec, Jure},
  booktitle={Temporal Graph Learning Workshop @ KDD},
  year={2025},
}

@article{wang2025hierarchical,
  title={Hierarchical Diffusion Policy: manipulation trajectory generation via contact guidance},
  author={Wang, Dexin and Liu, Chunsheng and Chang, Faliang and Xu, Yichen},
  journal={IEEE Transactions on Robotics},
  year={2025},
  publisher={IEEE}
}

@article{borse2026comfree,
  title={ComFree-Sim: A GPU-Parallelized Analytical Contact Physics Engine for Scalable Contact-Rich Robotics Simulation and Control},
  author={Borse, Chetan and Xie, Zhixian and Huang, Wei-Cheng and Jin, Wanxin},
  journal={arXiv preprint arXiv:2603.12185},
  year={2026}
}
\appendix
\section{APPENDIX}
\subsection{RL Observation Space Details}
\label{appendix:obs}
Here we provide detailed definition of three components in RL observation space.

\paragraph{Geometry component}
We downsample a set of $N$ keypoints $\mathcal{P} = \{\boldsymbol{p}_i\}_{i=0}^{N-1}$ from object point cloud,  defined in the \emph{object  frame}. These keypoints serve dual purposes: (i) as  an approximation for the object geometry $\mathcal{G}$, and (ii) as discrete candidates  for selecting contact location in    contact intention, i.e., $(\boldsymbol{\bar{c}}^1_k, \boldsymbol{\bar{c}}^2_k, ..., \boldsymbol{\bar{c}}^n_k)\subseteq \mathcal{P}$.

\paragraph{Target component}
To encode the target-aware observation, we define the \emph{keypoint goal flow} in the object frame, $\Delta \mathcal{P}_k = \{ \boldsymbol{\delta}_{i,k}|\boldsymbol{\delta}_{i,k} = \boldsymbol{p}_{i,k}^{\text{target}} - \boldsymbol{p}_i\}_{i=0}^{N-1}$, 
where $\boldsymbol{p}_{i,k}^{\text{target}}$ denotes the position of the same keypoint $\boldsymbol{p}_i$ when the object is at the  target, expressed in the current object frame at decision step $k$. The keypoint goal flow $\Delta \mathcal{P}_k$ represents the object-frame displacement of all  keypoints from the current object pose to the target. Encoding keypoint goals in the object frame is crucial for stable and geometry-consistent policy learning. Otherwise, the discrepency between body and world frames would force the policy to implicitly learn object translations and rotations, significantly increasing the learning burden.

\paragraph{Collision component}
To inform  contact  feasibility, we include \emph{per-keypoint clearance distances} as part of the observation. 
For each keypoint $\boldsymbol{p}_{i}$ at decision step $k$, we compute its minimum Euclidean distance $d_{i,k}$  to the environment (e.g., ground and walls). We collect these distances as $\mathcal{D}_k=\{d_{i,k}\}_{i=0}^{N-1}$.
$\mathcal{D}_k$ encode the \emph{contact feasibility} of each keypoint: keypoints with small clearance ($d_{i,k}\approx 0$) indicate it is in contact with their surrounding environment, thus are treated as invalid  candidates to be a contact location. Including $\mathcal{D}_k$ enables RL policy to avoid selecting contact locations that are physically blocked by the environment. For example, for a cube resting on a table, keypoints on the bottom face are  infeasible contact locations for the end-effectors.

\subsection{Details on MPC dynamics model}
\label{appendix:mpc_dyn}
We adopt the quasi-dynamic complementarity-free contact model from \cite{jin2024complementarity} for our relatively slow non-prehensile manipulation task, given by
\begin{equation}
\label{equ.comfree_dyn}
    \begin{aligned}
           & M\boldsymbol{v}^{+}
=   h\boldsymbol{\tau}(\boldsymbol{u})
\;+\;
 h \tilde{J}^{\top}\boldsymbol{\lambda},
 \\
 &\boldsymbol{\lambda} :=\mathrm{max}\Big(
- K\bigl(h^2\tilde{J} M^{-1}\boldsymbol{\tau}(\boldsymbol{u}) + \tilde{\boldsymbol{\phi}}\bigr),0\Big),
    \end{aligned}
\end{equation}
where $h$ is the time step and $\boldsymbol{v}^{+}$ is the next-step system velocity.  $M = \text{diag}(M_o,M_r)$ is the mass matrix, with object inertia $M_o$ and robot stiffness $M_r = h^2\text{diag}(K_r,\dots,K_r)$ for $n$ active contact points. The  force vector $\boldsymbol{\tau}(\boldsymbol{u}) = [\boldsymbol{\tau}_o, \boldsymbol{\tau}_r(\boldsymbol{u})]$ includes robot gravitational  force $\boldsymbol{\tau}_o$ and robot actuation force $\boldsymbol{\tau}_r(\boldsymbol{u})$. We assume robot is in  operational space impedance control and moves slowly. Thus, its damping and inertial effects are neglected, and end-effector force can be approximated as a spring model with stiffness $K_r$: $\boldsymbol{\tau}_r(\boldsymbol{u}) = K_r\boldsymbol{u}$. Here, the input $\boldsymbol{u}$ is desired end-effector displacement.  The contact-force vector $\boldsymbol{\lambda}$ is produced by a spring--damper model on the dual cones of the Coulomb friction cones, with $\tilde{J}$ and $\tilde{\boldsymbol{\phi}}$ the Jacobian and signed distances for all candidate dual cones, and $K$ the dual-cone stiffness.  Different 
contact modes, e.g., sticking, sliding, and separation, is captured by
different penetration regimes across facets of the dual cones.

With $\boldsymbol{v}^{+}$, the state after one-step forwarding of the contact dynamics is 
$
    \boldsymbol{x}^+ {=}  \boldsymbol{x} \oplus h\boldsymbol{v}^{+},
$
where $\oplus$ stands for the the integration  of system positions with velocities. The details of this closed-form contact model can be found in~\cite{jin2024complementarity}. 

\subsection{Implementation details of network backbones}
\label{appendix:backbone}
Two different network structures are used for the two options of subgoal parameterization in this work.
\begin{figure*}[!htbp]
    \centering
    \begin{subfigure}[b]{0.95\linewidth}
    \includegraphics[width=\linewidth]{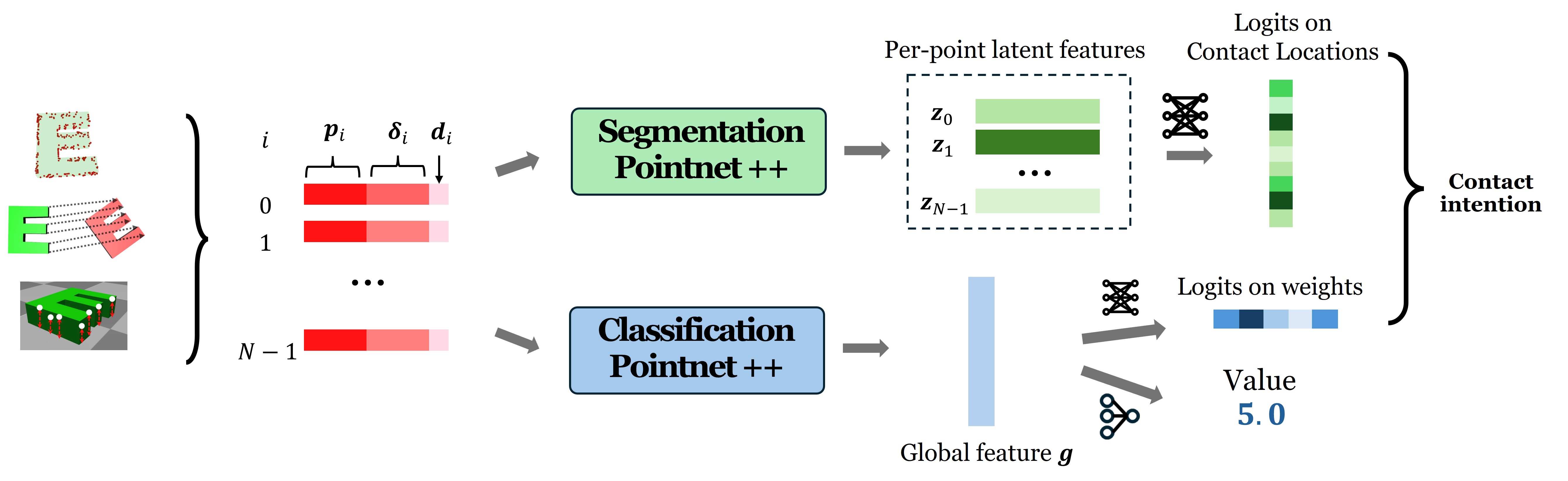}
    \caption{The dual-branch architecture design with pointnet++ backbone.}
    \end{subfigure}
    \begin{subfigure}[b]{0.95\linewidth}
    \includegraphics[width=\linewidth]{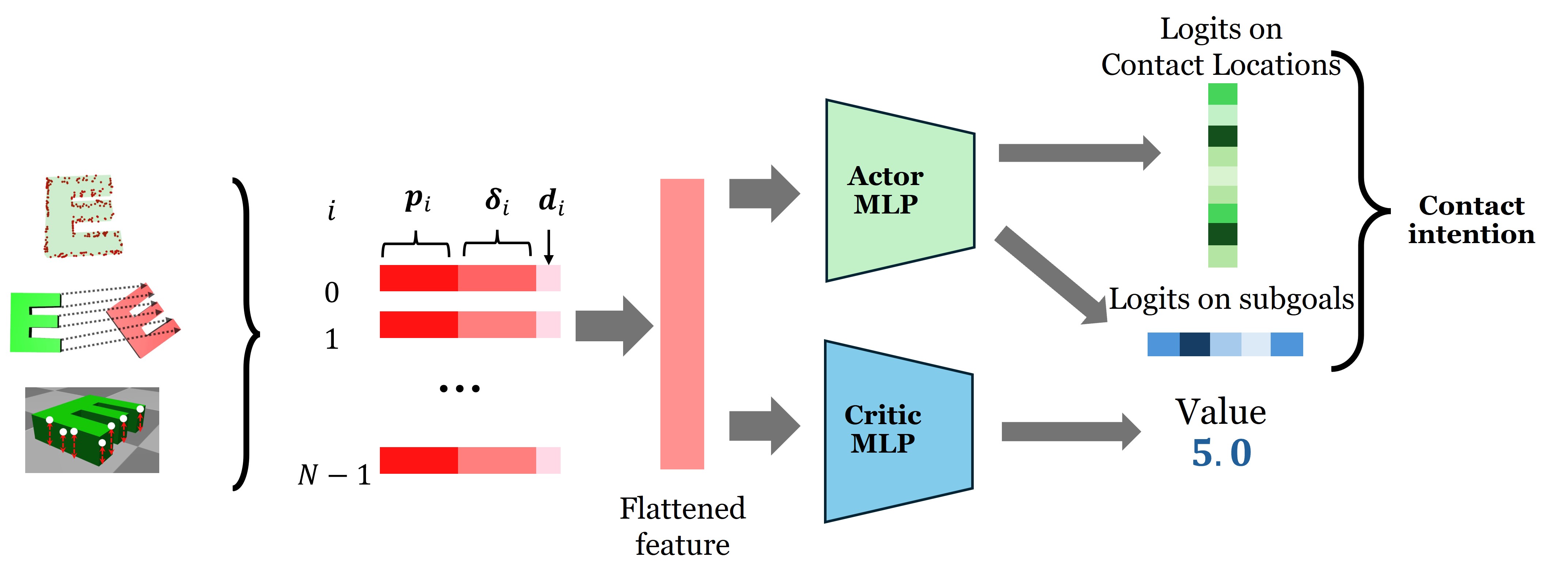}
    \caption{The policy architecture with MLP backbone.}
    \end{subfigure}
    \caption{RL policy architectures.}
    \vspace{-15pt}
    \label{fig.RL_policy}
\end{figure*}
\paragraph{Network structure with pointnet++ backbone} This framework (shown in Fig.~\ref{fig.RL_policy}(a)) adopts a \emph{dual-branch design} to process relatively dense per-keypoint features The input first goes through \textbf{geometric branch}: it processes per-keypoint features using a segmentation-style PointNet++ encoder to produce latent features $\boldsymbol{z}_i$, which are mapped to logits for contact location selection. Meantime, the input also goes through \textbf{global kinematic branch}, which aggregates all keypoint inputs into a single object-centric embedding $\boldsymbol{g}$ via a classification-style PointNet++ encoder. Linear heads then predict categorical logits over MPC cost weights $(\mathcal{W}_{\text{pos}}, \mathcal{W}_{\text{ori}})$ and a scalar value estimate for critic learning.  This architecture efficiently combines local geometric reasoning with global object-level context, enabling the policy to select feasible contact locations and implicitly specify object subgoals.

\paragraph{Network structure with MLP backbone} The MLP backbone provides a lightweight alternative for tasks with compact observations (shown in Fig.~\ref{fig.RL_policy}(b)) . It first flattens the input features into a single vector, which is then passed through shared fully connected layers. Separate prediction heads are used to output the contact points and subgoal selections, allowing the policy to decouple contact selection from subgoal selection. In addition, we use a separate value network with the same flattened input representation to estimate the state value for RL training.

\subsection{Detailed Task Setting in Sim and Real}
\label{appendix:task_setting}
\subsubsection{Robot Operational Space Control}
In the simulated experiments with full robot system, a Operational Space Controller (OSC) is implemented to transform the desired end-effector Cartesian space displacement output of Comfree-MPC to torque command in joint space. In the real-world testing with Franka robot arm, the robot is controlled by an OSC implemented in the panda-py library \cite{elsner2023taming}. The controller executes at 500~Hz, enabling fast reaction to the end-effector displacement command during manipulation.

\subsubsection{Robot Motion Planning for Contact Locations}
Between contacts, a heuristic is used to move the robot: the end-effector is lifted to disengage from the object and repositioned using OSC in free space. The OSC then guides the end-effector to the next intended contact location, where control switches back to the MPC for contact-rich interaction.

\subsubsection{Object Keypoints}
In geometry-generalized pushing, 256 keypoints are uniformly sampled from the vertical sides of all letters. In onject reorientation task, 256 keypoints are sampled uniformly from the cube surface. In environment-assisted object repositioning, we select a 3 by 3 grid from both top and bottom surface, forming a keypoint set with size 18.

\subsubsection{Distribution of Initial and Target Pose}

\begin{figure}[h]
  \centering
  \begin{subfigure}[b]{0.27\linewidth}
    \includegraphics[width=\linewidth]{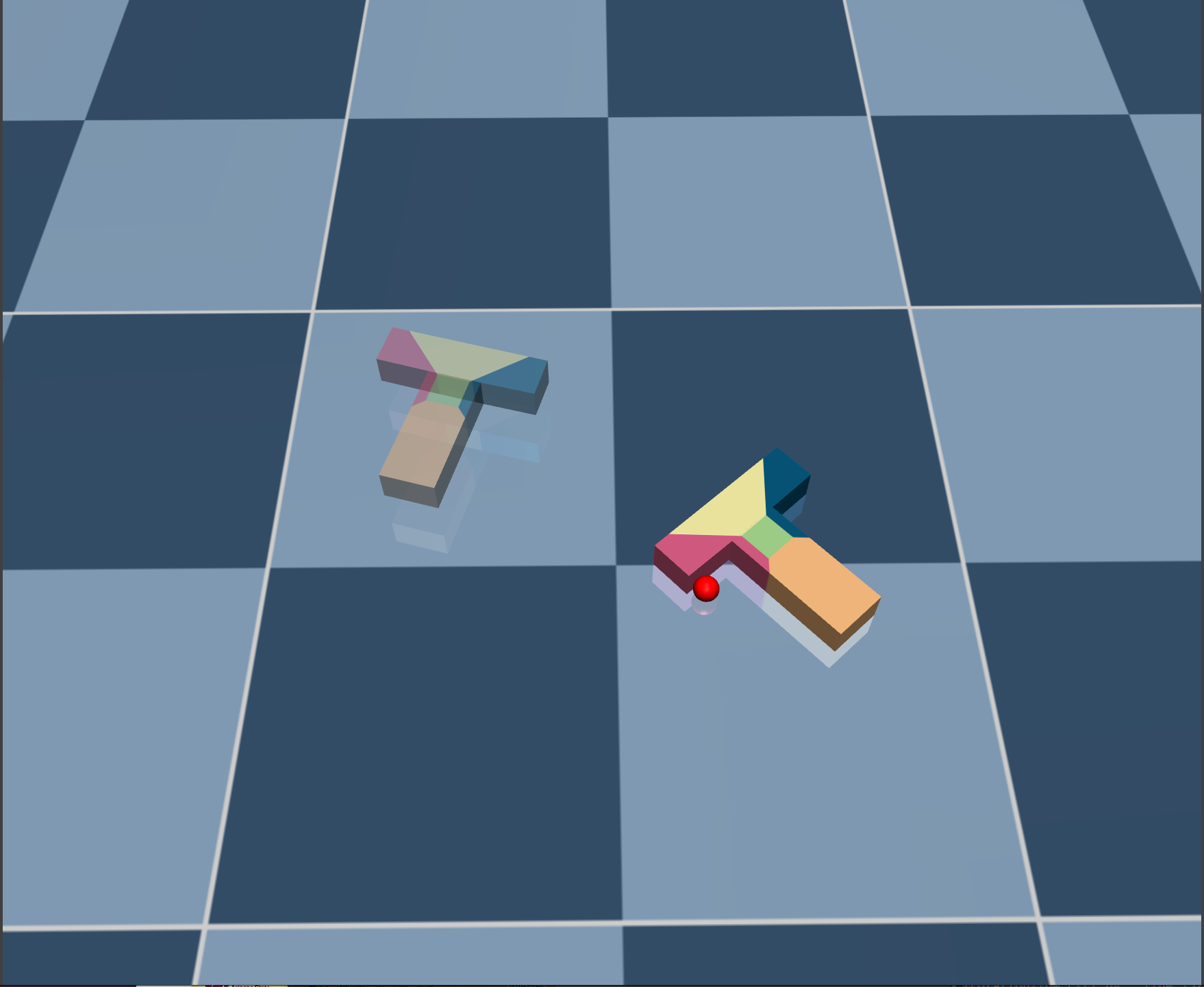}
  \end{subfigure}
   \hspace{10pt}
  \begin{subfigure}[b]{0.27\linewidth}
    \includegraphics[width=\linewidth]{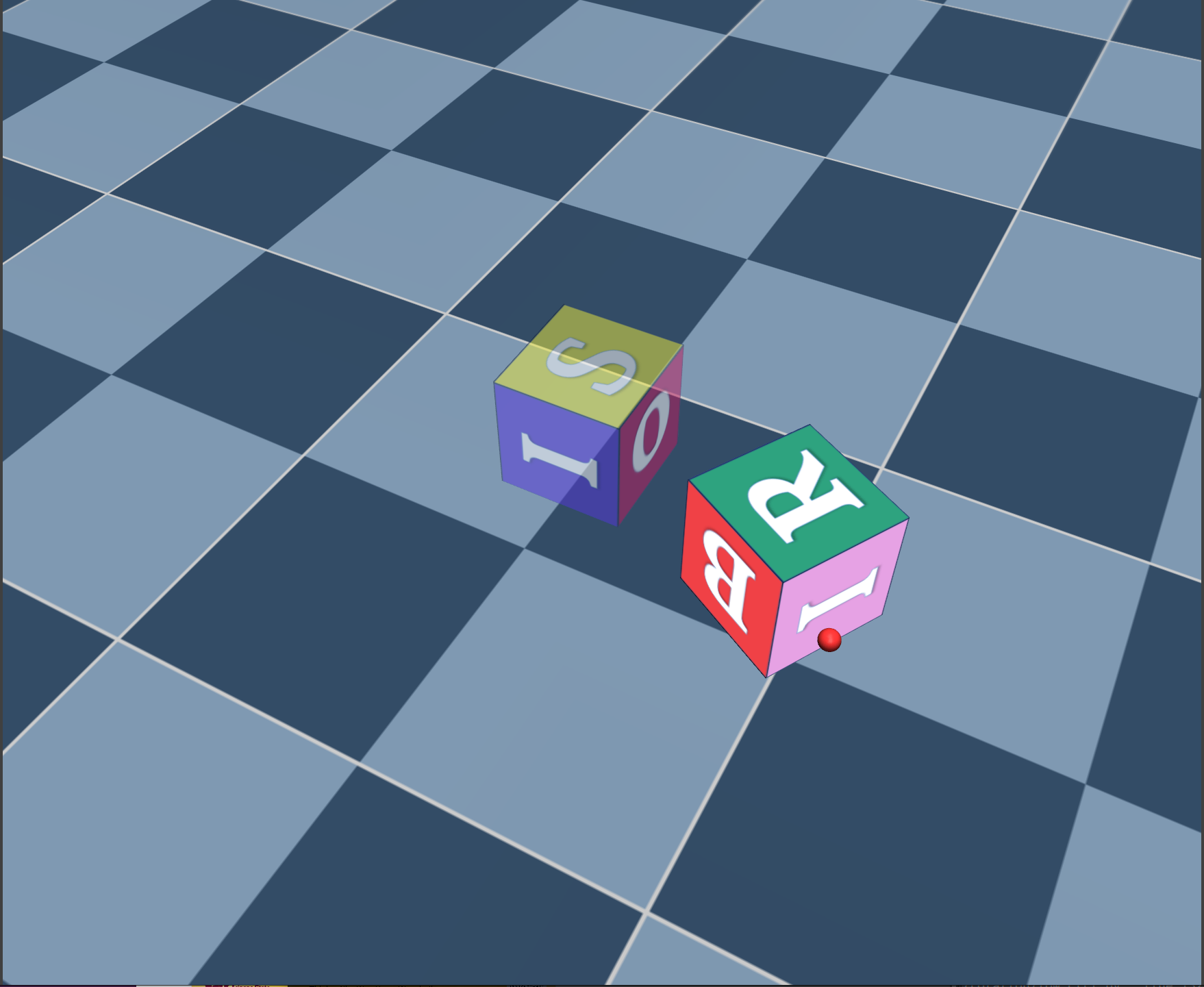}
  \end{subfigure}
  \hspace{10pt}
  \begin{subfigure}[b]{0.27\linewidth}
    \includegraphics[width=\linewidth]{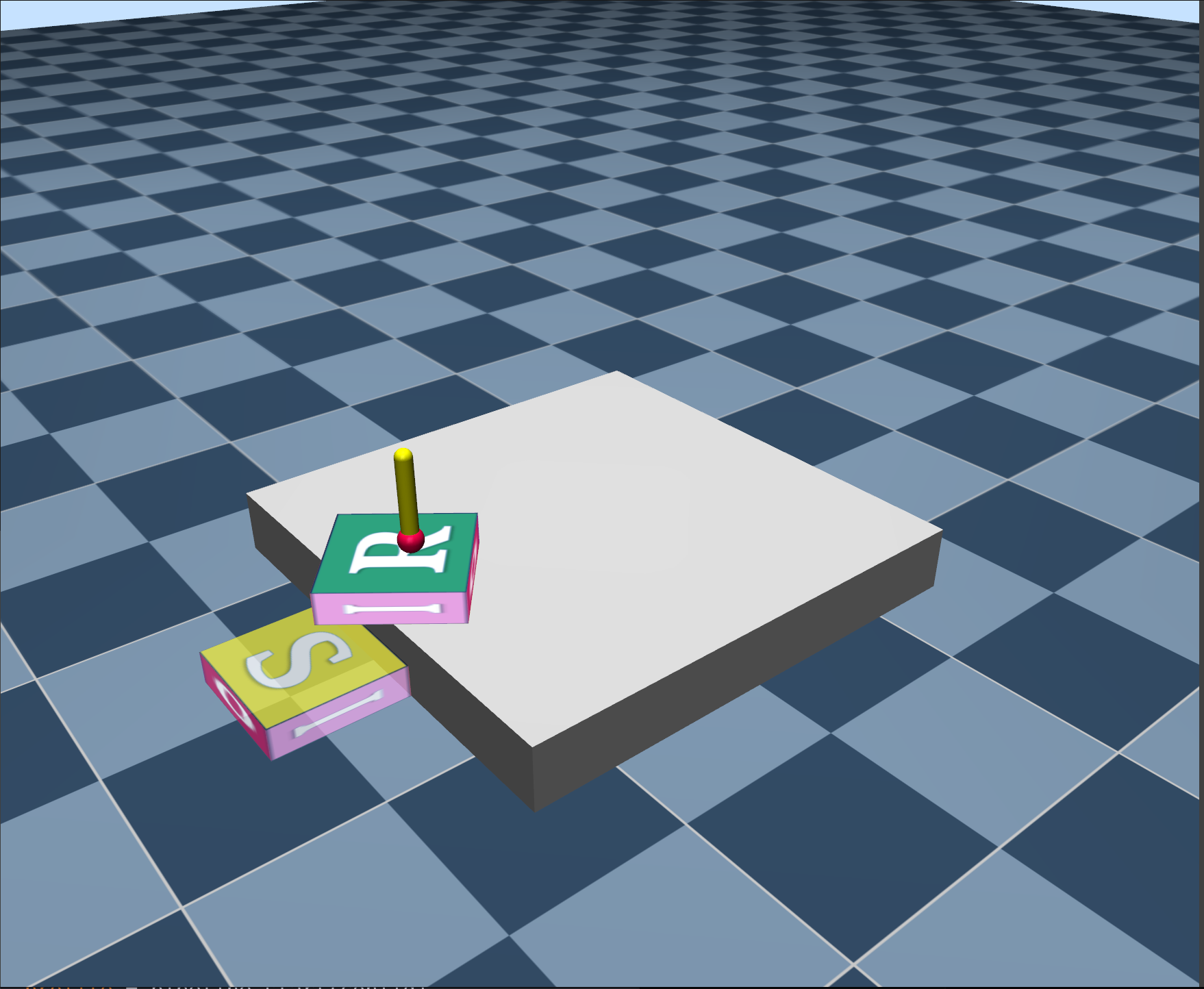}
  \end{subfigure}
  \caption{In the high-level RL policy training, an abstract end-effector  (red dot) replaces full  robot arm. The target  is  half-transparent. In simulated environment-assisted repositioning, we also attach a stick to the ball to simulate the collision behavior of the end-effector more precisely.}
  \vspace{-10pt}
  \label{fig.tasks}
\end{figure}

For each task, the initial and target translation of $xy$ coordinates are shown in Table \ref{tab:task_init_trans}. Example figures of the tasks are shown in \ref{fig.tasks}. Note that we use (P) to stand for Geometry-generalized pushing, (R) to stand for object reorientation, (E) to stand for  environment-assisted repositioning, and use Abstract, Full-Sim, Full-Real to stand for abstract end-effector setting, full robot setting in simulation, and real-robot setting. Additionally, each pushing target orientation is a planar rotation with an random $[-180^\circ, 180^\circ]$ angle about the $z$-axis, and each flipping target is defined a fixed $180^\circ$ rotation about the $x$-axis, and an additional rotation with a random $[-180^\circ, 180^\circ]$ angle about the z-axis. In environment-assisted repositioning task, the object starts from the fixed orientation of $(1,0,0,0)$ and fixed $z$ value of 0.1 (on the stair) to a fixed orientation target $(0,1,0,0)$, and fixed $z$ value of 0.02 (on the table).

\begin{table}[h!]
\centering
\caption{Initialization and target distributions for all tasks.}
\begin{tabular}{lcc}  
\toprule
\textbf{Task} & \textbf{Init Translation (x × y)} & \textbf{Target Translation} \\
\midrule
Abstract (P) & $[-0.25,0.25]^2$ & Same as init\\
Full-Sim (P) & $[0.4,0.6] \times [-0.1,0.1]$ & Same as init \\
Full-Real (P) & $[0.4,0.6] \times [-0.3,0.3]$ & $[0.4,0.5] \times [-0.1,0.1]$ \\
Abstract (R) & $[-0.1,0.1]^2$ & Same as init\\
Full-Sim (R) & $[0.4,0.6] \times [-0.1,0.1]$ & Same as init \\
Full-Real (R) & $[0.45,0.5] \times [-0.05,0.05]$ &  Same as init \\
Abstract (E) & $[-0.05,0.05] \times [0.1,0.3]$ & Fixed $(0.0,0.38)$\\
Full-Sim (E) & $[0.45,0.55] \times [-0.1,0.0]$ & Fixed $(0.5,0.08)$\\
Full-Real (E) & $[0.45,0.55] \times [-0.1,0.0]$ & Fixed $(0.5,0.08)$\\
\bottomrule
\end{tabular}
\label{tab:task_init_trans}
\end{table}

Success thresholds of all tasks are shown in Table \ref{tab:task_threshold}. The translational and orientational distance are shown in (9). The rotation distance is defined as $1 {-} \langle\boldsymbol{r}_H^{\text{obj}},\,\boldsymbol{\bar{r}}^{\text{target}}\rangle^2$.
A trial is considered successful if the object satisfies both thresholds. Note that in realworld experiments of object reorientation and  environment-assisted repositioning, the threshold is slightly loosened to compensate for tracking error.

\begin{table}[h!]
\centering
\caption{Success Threshold for all tasks.}
\begin{tabular}{lcc}  
\toprule
\textbf{Task} & \textbf{Trans Threshold (m)} & \textbf{Rot Threshold} \\
\midrule
Abstract (P) & 0.05 & 0.03\\
Full-Sim (P) & 0.05 & 0.03 \\
Full-Real (P) & 0.03 & 0.02 \\
Abstract (R) & 0.05 & 0.03\\
Full-Sim (R) & 0.05  & 0.03 \\
Full-Real (R) & 0.075 &  0.03\\
Abstract (E) & 0.08 & 0.03\\
Full-SIm (E) & 0.08 & 0.03\\
Full-Real (E) & 0.085 &  0.03\\
\bottomrule
\end{tabular}
\label{tab:task_threshold}
\end{table}

\subsubsection{Subgoal Parameterization}
Both the geometry-generalized pushing and pivoting-flipping based object reorientation use option 1 in Section \ref{sec.rl_action} to parameterize the contact-intention subgoals, and use pointnet++ backbone as RL network structure. The environment-assisted object repositioning uses option 2 and MLP backbone.

\subsection{Settings of RL policy, reward and training}
\label{appendix:rl_setting}
\subsubsection{RL Policy Setting}
\paragraph{Pointnet++ Setting in RL policy} 
We employ two PointNet++ feature extractors---a \emph{Geometric Processing Branch} for dense per-point reasoning and a \emph{Global Kinematic Processing Branch} for object-level aggregation.  The Geometric Processing Branch follows a segmentation-style PointNet++ architecture, the Global Kinematic Processing Branch follows a classification-style PointNet++ architecture. Those two branches shares the same Pointnet++ setting, as shown in Table~\ref{tab:pn}. We use the implementation of Pointnet++ networks in pytorch-geometric library \cite{pyg}.  Specifically, although the nominal input channel number is 4 (per-point goal flow and contact-distance features), the 3D keypoint coordinates themselves are normalized and concatenated to this input inside the module.

\paragraph{MLP backbone setting}
In MLP backbone, the policy network is with width 64 and 2 hidden layers. Seperate heads are used to select contact point and subgoals. The value network is also with width 64 and 2 hidden layers. All networks uses Tanh for activation.

\begin{table}[h]
\centering
\caption{PointNet++ hyperparameters.}
\label{tab:pn}
\begin{tabular}{ll}
\toprule
\textbf{Parameter} & \textbf{Value} \\
\midrule
Input channels                    & 4 \\
Output feature dimension          & 128 \\
Dropout                           & 0.0 \\
Farthest point sampling random start & True \\
Include point coordinates in Input  & True \\
Position normalization            & True \\
Activation                        & LeakyReLU \\
\bottomrule
\end{tabular}
\end{table}

\paragraph{MPC weights action space} 
When using method 1 for subgoal parameterization, we choose $\mathcal{W}_{\text{pos}} = \{0, 50, 100, 150, 200\}$ and $\mathcal{W}_{\text{ori}} = \{0, 2, 4, 6, 8\}$.

\paragraph{Discrete subgoal action space} 
When using Method 2 for subgoal parameterization, the RL agent selects from a discrete set of pose candidates defined by the Cartesian product of the following choices: (i) the $x$-offset in the object body frame, chosen from $\{-0.03, 0, 0.03\}$; (ii) the $y$-offset in the object body frame, chosen from $\{-0.03, 0, 0.03\}$; (iii) the absolute $z$ position, chosen from $\{0.02, 0.10\}$; and (iv) the absolute orientation, represented as a quaternion, chosen from $\{(1,0,0,0), (0.707,-0.707,0,0), (0,1,0,0)\}$.

\subsubsection{Scaling of Observation}
To improve training stability and generalization across objects of different sizes, all observation components are normalized (divided) by the characteristic size of the object. We set this size to be 0.15 for all letters, 0.10 for the cube and 0.16 for the rectangular object in three tasks.

\subsubsection{RL Reward Setting}
\label{appendix:rl_reward}
We train the high-level RL policy using a reward
\begin{equation}
\begin{aligned}
r_k = \big(w_1 r_{\text{dense}}  + w_2 r_{\text{target}}\big)(1+r_{\text{feasible}}) + w_3 r_{\text{feasible}},
\end{aligned}
\end{equation}
where $w_1,w_2,w_3$ are the weights for the negative dense keypoint flow reward ($r_{\text{dense}} {=}-  \frac{1}{N} \sum_{\boldsymbol{\delta}_{i,k} \in \Delta \mathcal{P}_k} \lVert \boldsymbol{\delta}_{i,k} \rVert$), sparse target-reaching reward ($r_{\text{target}}{=}1$ at final target), and negative contact feasibility reward $r_{\text{feasible}}{=}-1$ when the contact location is  invalid  (e.g., contact location on the bottom surface); otherwise $r_{\text{feasible}}=0$.

The reward hyperparameter is set in Table $\ref{tab:reward}$. We also add a -1 penalty in  environment-assisted repositioning for object leaving the $xy$ workspace $(-0.2,0.2)\times(-0.1,0.75)$. Specifically, to calculate the term $\frac{1}{N} \sum_{\boldsymbol{\delta}_{i,k} \in \Delta \mathcal{P}_k} \lVert \boldsymbol{\delta}_{i,k} \rVert$, we use the target component after scaling.
\begin{table}[!htbp]
    \centering
    \caption{Task-Specific Reward Parameters}
    \label{tab:reward}
    \begin{tabular}{lccc}
    \toprule
        \textbf{Param} & \textbf{Geometry-Generalized Pushing} & \textbf{Object Reorientation} & \textbf{Env-contact Task} \\
        \midrule
        $w_1$ & 0.1 & 0.1 & 0.1\\
        $w_2$ & 5.0 & 5.0 & 50.0\\
        $w_3$ & 0 & 1.0 & 1.0\\
        \bottomrule
    \end{tabular}
\end{table}

\subsubsection{RL Training Setting}
All reinforcement-learning experiments use Proximal Policy Optimization (PPO) in Stable-Baselines3 \cite{stable-baselines3} with the hyperparameters summarized in Table~\ref{tab:ppo-hparams}.  The exceptions are: (1) $n_{\text{steps}}$ in end-to-end model baseline comparison, which is set to be 128 to match its higher decision frequency. (2) when using MLP backbone, the clipping range is set to 0.2 and the GAE parameter is set to 0.96.
 All hyperparameters are carefully tuned to ensure stable optimization and strong empirical performance across tasks. The proposed policy outputs a structured action $\boldsymbol{a}_k$ composed of multiple discrete components, corresponding to different aspects of contact intention and MPC parameterization. During training, the log-probability and entropy terms used in the PPO objective are computed as sums over individual components, enabling balanced policy update.
\begin{table}[h!]
\centering
\caption{PPO hyperparameters used across all experiments.}
\begin{tabular}{l c}
\hline
\textbf{Parameter} & \textbf{Value} \\
\hline
Number of environments ($n_{\text{envs}}$) & 32 \\
Rollout length ($n_{\text{steps}}$) & 32 \\
Optimization epochs ($n_{\text{epochs}}$) & 5 \\
Batch size & 512 \\
Clipping range ($\epsilon$) & 0.3 \\
Discount factor ($\gamma$) & 0.99 \\
GAE parameter ($\lambda$) & 0.95 \\
Entropy coefficient ($c_{\text{ent}}$) & 0.0 \\
Value loss coefficient ($c_{\text{vf}}$) & 0.5 \\
\hline
\end{tabular}
\label{tab:ppo-hparams}
\end{table}

\paragraph{Tempertaure Scaling of Logits}
During training when using pointnet++ backbone, a temperature scaling of $0.1$ (i.e., scaling the logits by a factor of $10$) is applied to the contact-selection logits to sharpen the resulting categorical distribution and improve learning efficiency, while the logits corresponding to the MPC cost weights are used without additional scaling. After the scaling process, a softmax function is applied on logits to form the discrete probability distribution among contact point/ MPC weights selection during training.

\paragraph{Learning-Rate Scheduling.}
When using the pointnet++ backbone, we employ a monotonically decaying learning-rate schedule of the form
\[
\text{LR}(p) = 10^{-4} \cdot \beta^{(p - 1)}, \qquad p \in [1, 0],
\]
where $p$ denotes normalized remaining training progress from start ($p{=}1$) to end ($p{=}0$).  
For the main experiments, we set $\beta = 100$. In the  baseline comparisons, we use a smaller decay factor $\beta = 10$ on end-to-end and HACMan style models to enable stronger updates. In ablation studies, we still set $\beta = 100$. When using MLP backbone, we use fixed learning rate of 0.003.

\paragraph{Domain Randomization}
We apply domain randomization during training to improve robustness to model mismatch and real-world variations. Specifically, we randomize the object mass and friction coefficient, as well as the low-level actuation gains. The object mass is scaled within $[0.75, 1.25]$, and the friction coefficient is scaled within $[0.5, 1.5]$. The actuation proportional gain $K_p$ and derivative gain $K_d$ are sampled from $[80.0, 120.0]$ and $[0.0, 10.0]$, respectively. The full randomization ranges are summarized in Table~\ref{tab:domain_randomization}.

\paragraph{Parallelization}
To improve training and execution efficiency, we parallelize both simulation and optimization components. MuJoCo rollouts are executed with multi-threading to collect environment interactions in parallel, while the MPC is also parallelized across multiple solver instances. This allows parallel training for RL.

\begin{table}[h]
\centering
\caption{Domain randomization ranges used during training.}
\label{tab:domain_randomization}
\small
\begin{tabular}{lc}
\toprule
\textbf{Parameter} & \textbf{Range} \\
\midrule
Mass scale & $[0.75, 1.25]$ \\
Friction scale & $[0.5, 1.5]$ \\
Actuation $K_p$ & $[80.0, 120.0]$ \\
Actuation $K_d$ & $[0.0, 10.0]$ \\
\bottomrule
\end{tabular}
\end{table}

\subsection{MPC Setting in Sim and Real}
\label{appendix:mpc_setting}
Our low-level MPC is implemented via CasADi(\cite{CasADi}), using IPOPT (\cite{wachter2006implementation}) as optimization algorithm. While the core MPC formulation is shared across all experiments, physical parameters are adjusted differently for sim/real task to reflect their distinct physical properties.  
Table~\ref{tab:task_params} summarizes these task-dependent settings. Among those parameters, $M$ in \eqref{equ.comfree_dyn} is formed by:
\begin{equation}
    M = \textbf{diag}(m,m,m,\mathbf{i},\mathbf{i},\mathbf{i},h^2K_r,h^2K_r,h^2K_r)
\end{equation}
to match the 9 DOFs of the system with the abstract end-effector (6 for object, 3 for robot). In the equation above, $m$ stands for object mass, $\mathbf{i}$ stands for the rotation inertia, $K_r$ is the stiffness of the robot end-effector. Note that these parameters do not necessarily correspond to the true physical values of the object or the robot in the environment. Rather than aiming for exact physical fidelity, these parameters are tuned to provide a stable and effective predictive model for real-time planning. $K,h$ along with $M$ are then used in \eqref{equ.comfree_dyn}. $H, w_c$ and the control range scale is used to parameterize the optimization problem \eqref{equ.mpc}. $\mu$ is used in the calculation of the Jacobian of the dual cones $\tilde{J}$ in \eqref{equ.comfree_dyn}. $T$ denotes the number of environment steps, or MPC updates, executed for each RL command. In simulation, $T$ corresponds to a fixed number of steps, whereas in real-world experiments it specifies the total MPC execution time due to the asynchronous control stack. MPC parameter is adjusted to fit the physical properties of the system.  Importantly, these adjustments are lightweight, physically intuitive, and do \emph{not} require retraining the RL policy, as the policy reasons only on contact intention. For details of how to form the dynamics \eqref{equ.comfree_dyn} and MPC optimization problems \eqref{equ.mpc} using those parameters, please refer to \cite{jin2024complementarity}.

\begin{table}[h]
\centering
\caption{Task-specific MPC parameter settings.}
\label{tab:task_params}
\begin{tabular}{lcccccc}
\toprule
\textbf{Param} & \textbf{Sim-P} & \textbf{Real-P} & \textbf{Sim-R} & \textbf{Real-R} & \textbf{Sim-E} & \textbf{Real-E}\\
\midrule
Dynamics time step $h$ & 0.05 & 0.05 & 0.05 & 0.05 & 0.05 & 0.05 \\
Robot stiffness $K_r$ & 0.01 & 0.01 & 0.01 & 0.01 & 0.1 & 0.1 \\
MPC horizon $H$ & 3 & 3 & 3 & 3 & 3 & 3 \\
Contact weight $w_c$ & 0.2 & 0.05 & 0.2 & 0.2 & 0.2 & 0.2 \\
Object inertia $\mathbf{i}$ & 0.0005 & 0.0001 & 0.0005 & 0.001 & 0.0005 & 0.0005 \\
Object mass $m$ & 0.2 & 0.05 & 0.2 & 0.1 & 0.2 & 0.2 \\
Control range scale & 0.03 & 0.015 & 0.03 & 0.02 & 0.03 & 0.04 \\
Stiffness coefficient $K$ & 0.1 & 0.15 & 0.1 & 0.15 & 3.0 & 2.0 \\
Friction coefficient $\mu$ & 1.0 & 0.5 & 1.0 & 1.0 & 2.0 & 1.0 \\
Ctrl Steps per RL cmd $T$ & 15 & 5(s) & 15 & 5(s) & 15 & 1(s) \\
\bottomrule
\end{tabular}
\end{table}

\subsection{Detailed Settings in End-to-End Comparison}
\label{appendix:end2end}
For simplicity, this comparison is conducted for the pushing task only with letter T. The task setting is the same as the geometry-generalized pushing task, while the only object used in training and evaluation is "\textbf{T}". For the action space, the end effector is constrained to move in the $xy$-plane with a fixed $z$ position,  with a range $[-0.05, 0.05]^2$ representing delta $xy$ motions. The observation includes object keypoints, goal flow, per-point clearance, and end-effector flow, defined as the distance vectors from the end effector to all keypoints. All components are expressed in the world frame to remain consistent with the predicted $xy$ motion, which is also defined in the world frame.

We found that directly using the reward   in Appendix~\ref{appendix:rl_reward} leads to the struggling of the  end-to-end policy to effectively explore the effect contact with the object. To facilitate object interaction, we add an additional reward term
$
    0.5 \exp\!\left(-{\|\boldsymbol{d}_{\text{obj,ee}}\|^2}/{0.02}\right),
$
to encourage the end effector to approach the object, where $\boldsymbol{d}_{\text{obj,ee}}$ denotes the distance between the end effector and the object center. Episodes are also terminated early when $\|\boldsymbol{d}_{\text{obj,ee}}\| \ge 0.25$, with a penalty of $-5$, to discourage drifting away from the object.

\vspace{-10pt}
\subsection{Realworld Experiment Setup}
\vspace{-10pt}
\begin{wrapfigure}{r}{0.45\textwidth}
    \centering
    \vspace{-10pt}
    \includegraphics[width=0.43\textwidth]{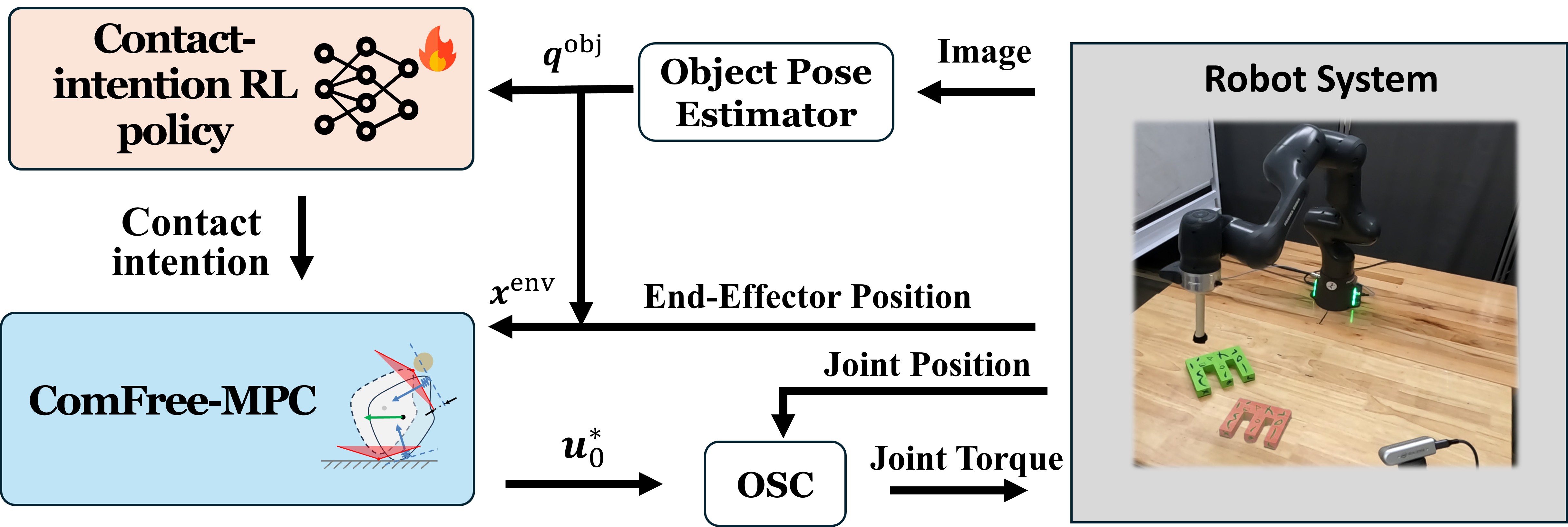}
    \caption{The Diagram of real-world system.}
    \label{fig.real}
    \vspace{-10pt}
\end{wrapfigure}
\label{appendix:realworld_setup}
We deploy the  high-level policy trained in simulation in Section \ref{sec.experiment} on  real-world  system.  Fig. \ref{fig.real} shows the hardware setup.  The  system composed of four modules. (1) The high-level RL policy module. We use the stochastic version of the trained  RL policy for contact-point selection in object reorientation task. This helps prevent the system from getting stuck due to noisy pose estimation.  (2) Low-level ComFree-MPC module ($\sim$100Hz). (3) Robot operational space control (OSC) stack ($\sim$500Hz) \cite{elsner2023taming}. It receives the end-effector displacement command from ComFree-MPC. (4) The vision perception module  ($\sim$40Hz), which uses the TwinTrack \cite{yang2025twintrack} and FoundationPose++ \cite{wen2024foundationpose,Wenhao_Yan_and_Jie_Chu_FoundationPose_2025}.  
These components interact through  LCM communication middleware\cite{huang2010lcm}. All objects are 3D-printed from the exact STL meshes used during training. Full result for each letter in pushing, object reorientation and environment-assisted object repositioning is shown in Table \ref{tbl.real_result_full}.

\begin{table}[]
\centering
\vspace{-10pt}
\caption{Result in real-world non-prehensile manipulation}
\label{tbl.real_result_full}
\begin{tabular}{lcc}
\toprule
\textbf{Object} & \textbf{Success rate} & \textbf{Num. steps}\\
\midrule
Push-E & 100.0\% & 5.60 $\pm$ 2.72 \\
Push-F & 100.0\% & 8.70 $\pm$ 6.53 \\
Push-H & 100.0\% & 5.20 $\pm$ 2.04 \\
Push-I & 70.0\% & 7.14 $\pm$ 2.91 \\
Push-K & 100.0\% & 6.20 $\pm$ 3.08 \\
Push-L & 100.0\% & 9.60 $\pm$ 3.89 \\
Push-N & 100.0\% & 6.20 $\pm$ 3.85 \\
Push-T & 100.0\% & 4.70 $\pm$ 2.71 \\
Push-V & 100.0\% & 6.40 $\pm$ 4.33 \\
Push-X & 100.0\% & 5.30 $\pm$ 3.06 \\
Push-Y & 100.0\% & 9.00 $\pm$ 3.92 \\
Push-Z & 100.0\% & 6.60 $\pm$ 6.24 \\
3D Reorientation & 100.0\% & 24.96 $\pm$ 15.5 \\
Env-contact task & 88.0\% & 5.92 $\pm$ 4.35 \\
\bottomrule
\end{tabular}
\vspace{-5pt}
\end{table}

\subsection{Baseline Comparison in Real-World Setting}
\label{appendix:realworld_baseline}
We evaluate our method against baseline approaches in real-world experiments, to demonstrate the advantage of the proposed hierarchical approach on sim-to-real generalization and long-horizon manipulation performance. 

\begin{figure}[t]
  \centering
  \begin{subfigure}[b]{0.9\linewidth}
    \includegraphics[width=\linewidth]{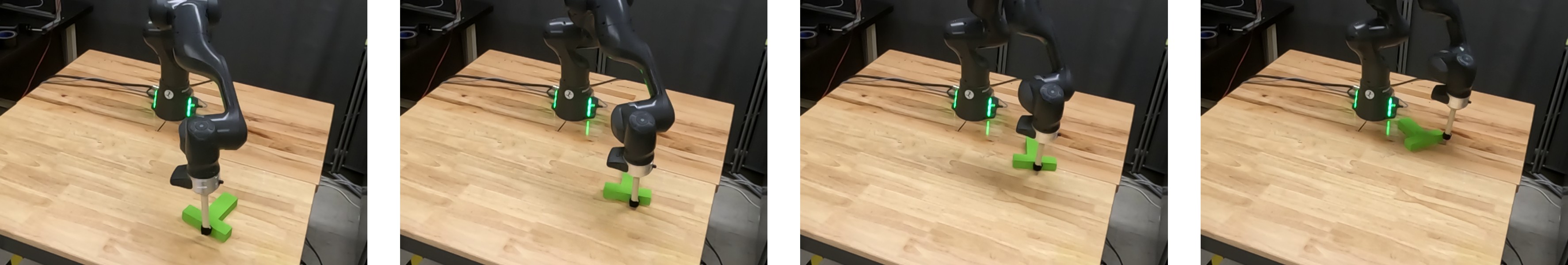}
    \caption{End2end: Excessive pushing of the object.}
    \label{fig.fail_modes_1}
  \end{subfigure}\\
  \begin{subfigure}[b]{0.9\linewidth}
    \includegraphics[width=\linewidth]{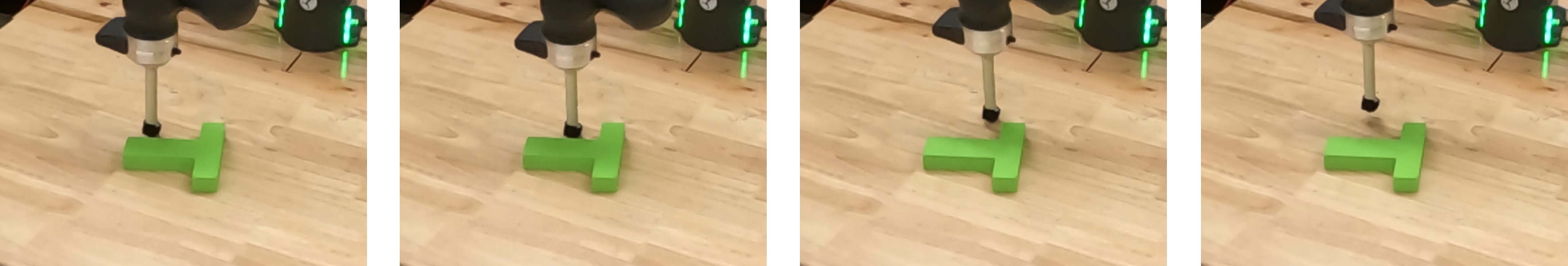}
    \caption{End2end: failure to establish contact.}
    \label{fig.fail_modes_2}
  \end{subfigure}\\
  \begin{subfigure}[b]{0.9\linewidth}
    \includegraphics[width=\linewidth]{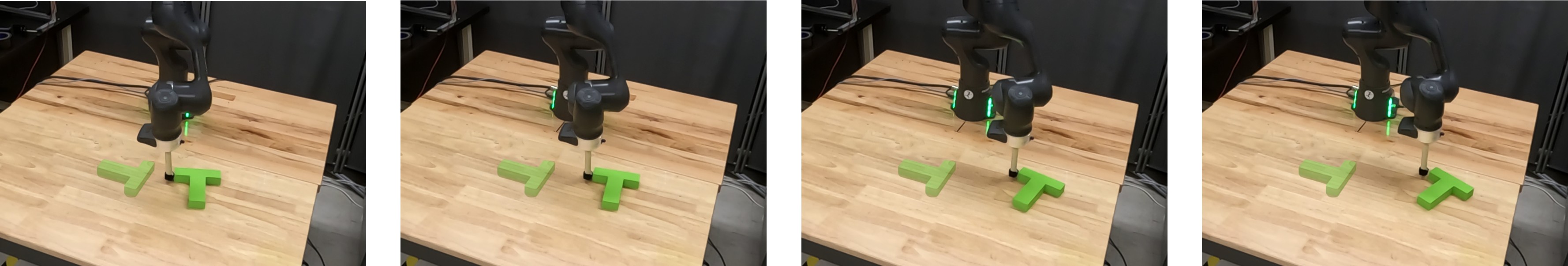}
    \caption{MPC only: stuck end-effector.}
    \label{fig.fail_modes_3}
  \end{subfigure}
  \caption{Different failure modes in real-world baseline experiments.}
  \vspace{-15pt}
  \label{fig.fail_modes}
\end{figure}

\paragraph{End-to-end RL policy}
We evaluate the real-world performance of the end-to-end RL policy by replacing the hierarchical policy in Fig.~\ref{fig.real} with the policy trained in Sec.~\ref{sec:sim_baseline}. The end-effector $z$-coordinate is fixed at $0.02$m, while the desired $\Delta xy$ position is provided by the planar policy output and tracked using an OSC controller. The policy is tested on the pushing task only on letter T, consistent with its simulation training setup. The success rate is $35\%$, with two primary failure modes: excessive pushing that causes the object to move out of reach or leads to irregular robot configurations (see Fig. \ref{fig.fail_modes_1}), and failure to establish contact due to perception errors in estimating object pose (see Fig. \ref{fig.fail_modes_2}). In contrast, the proposed hierarchical policy achieves $100\%$ success under the same setting. This result highlights the superior sim-to-real generalization capability of our approach.

\paragraph{MPC-only policy}
We also evaluate the real-world performance of an MPC-only policy, where the robot is controlled solely by the comfree-MPC without guidance from a learned high-level policy. The contact target is set to the mass center of the object mesh with a fixed set of terminal cost weights. The policy is tested on the push-T task under the same setup. While MPC provides stable low-level control, the absence of high-level contact intention leads to direct failure, particularly in selecting effective contact locations. As a result, the policy often exhibits inefficient or myopic behavior. In practice, the end-effector may initially push the object but then get stuck mid-execution (see Fig. \ref{fig.fail_modes_3}), as the optimizer struggles to simultaneously satisfy contact, positional, and rotational objectives. Compared to the $100\%$ success rate of the proposed hierarchical method, this highlights the importance of integrating learned high-level guidance with model-based control for long-horizon, contact-rich manipulation.

\subsection{Real-World Policy Rollout Visualizations}
\label{appendix:realworld_fig}
\begin{figure}[!htbp]
  \centering
  \begin{subfigure}[b]{0.9\linewidth}
    \includegraphics[width=\linewidth]{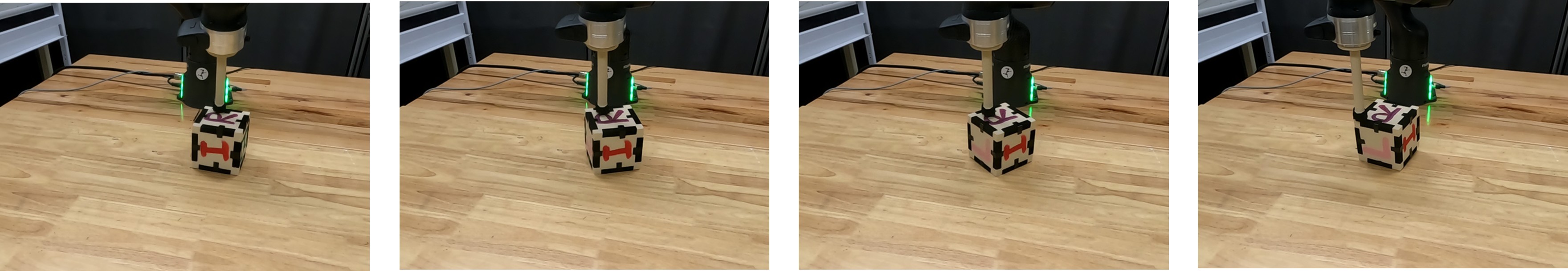}
    \caption{Reorientation by sliding on the top.}
  \end{subfigure}\\
  \begin{subfigure}[b]{0.9\linewidth}
    \includegraphics[width=\linewidth]{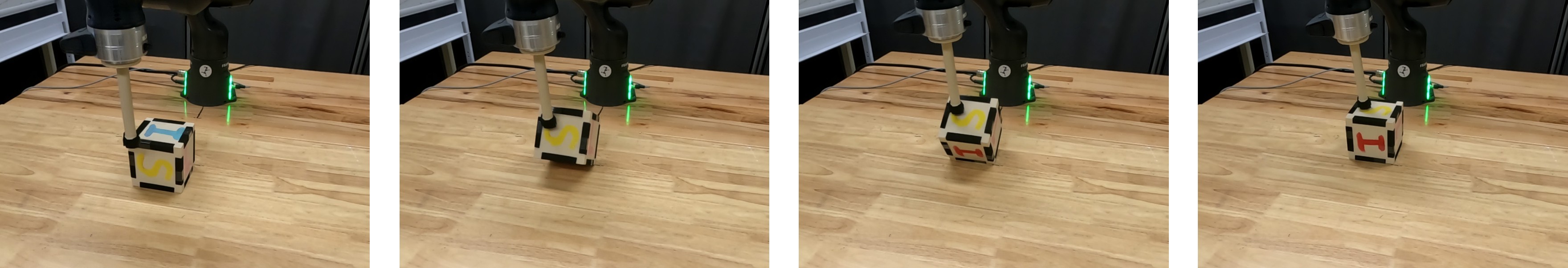}
    \caption{Edge pivoting.}
  \end{subfigure}\\
  \begin{subfigure}[b]{0.9\linewidth}
    \includegraphics[width=\linewidth]{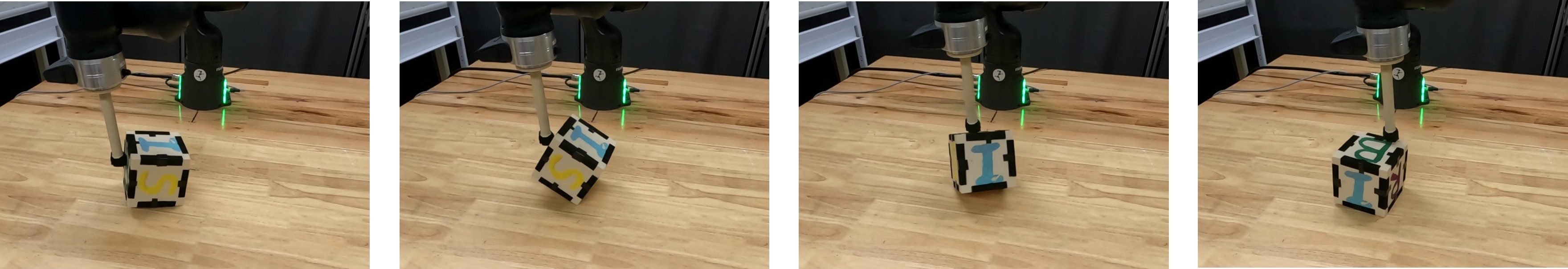}
    \caption{Corner pivoting.}
  \end{subfigure}
  \caption{Different contact strategies in the 3D reorientation experiment. Those different strategies in (a) (b) (c) are observed in separate trials.}
  \label{fig.flip_modes}
\end{figure}

\begin{figure}[!htbp]
  \centering
  \begin{subfigure}[b]{0.9\linewidth}
    \includegraphics[width=\linewidth]{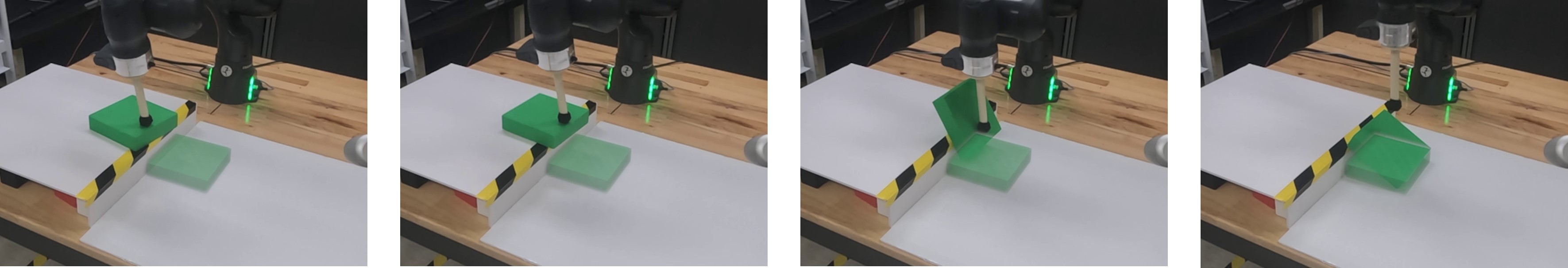}
  \end{subfigure}\\
  \vspace{5pt}
  \begin{subfigure}[b]{0.9\linewidth}
    \includegraphics[width=\linewidth]{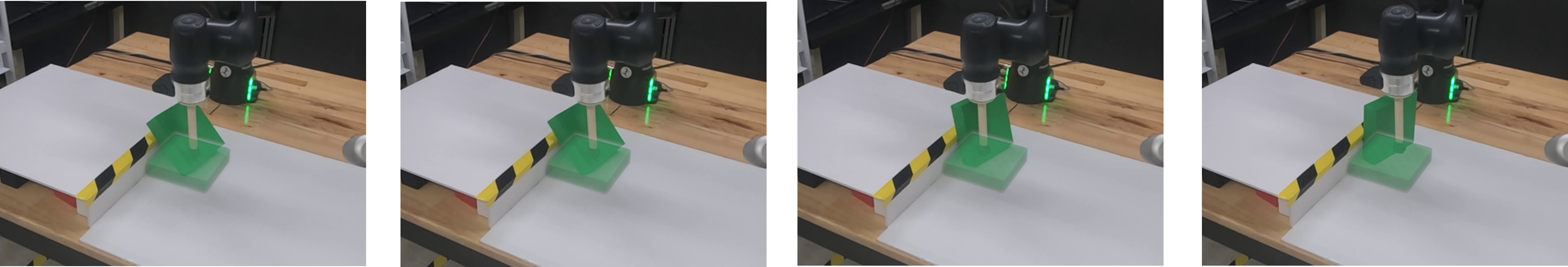}
  \end{subfigure}\\
  \vspace{5pt}
  \begin{subfigure}[b]{0.9\linewidth}
    \includegraphics[width=\linewidth]{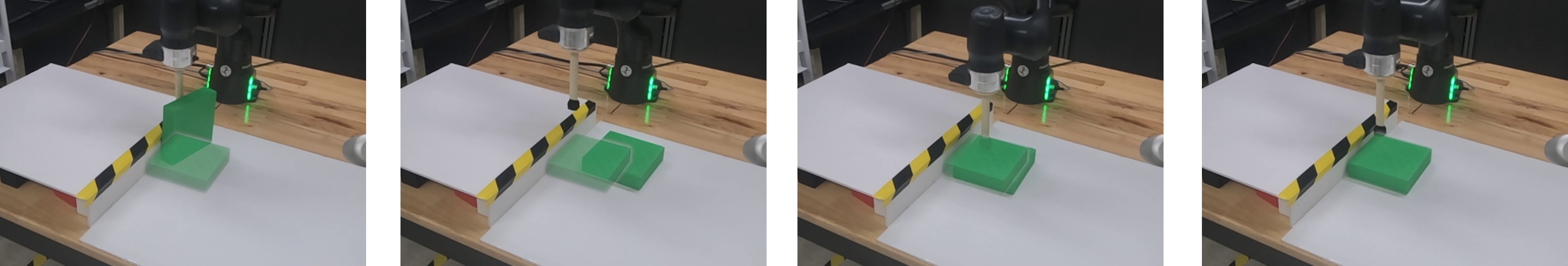}
  \end{subfigure}
  \caption{An example run of real-world environment-assisted object reposition task. Top:the robot slides on the top to make object near edge of stair, leveraging the table edge as an environmental support to pivot the object upward toward a vertical pose. Mid: The robot shifts to a bottom nudge that stabilizes the object from incline to the desired upright configuration. Last row: the robot shift the contact location and leverages the stair edge to topple the object toward the final flip pose.}
  \vspace{-15pt}
  \label{fig.edge_real}
\end{figure}

Fig. \ref{fig.flip_modes} shows diverse contact interaction strategies in object reorientation task, Fig. \ref{fig.edge_real} shows an example run of  environment-assisted repositioning.

\end{document}